\documentclass[english, 10pt, twocolumn, twoside]{IEEEtran}
\usepackage{amsmath}
\usepackage{amssymb}
\usepackage[english]{babel}
\usepackage{slashbox}
\usepackage{times}
\usepackage{epsfig}
\usepackage{graphicx}
\usepackage{diagbox}
\usepackage{multirow}
\usepackage{color}
\usepackage{url}

\newcommand{\eg}{\emph{e.g.}}
\newcommand{\etal}{\emph{et al.}}
\newcommand{\ie}{\emph{i.e.}}
\newcommand{\tabincell}[2]{\begin{tabular}{@{}#1@{}}#2\end{tabular}}
\newcounter{RomanNumber}

\DeclareMathOperator*{\argmin}{arg\,min}

  \usepackage{cite}
\ifCLASSINFOpdf
\else
\fi
\begin{document}
%
\title{Learning to Reconstruct and Understand Indoor Scenes from Sparse Views}
%
%
%
%

\author{Jingyu Yang,~\IEEEmembership{Senior Member,~IEEE,}~Ji Xu,~Kun Li,~\IEEEmembership{Member,~IEEE,}
        ~Yu-Kun Lai,~\IEEEmembership{Member,~IEEE,}\\Huanjing Yue,~\IEEEmembership{Member,~IEEE,}~Jianzhi Lu,~Hao Wu, and Yebin Liu,~\IEEEmembership{Member,~IEEE}

\thanks{
 Jingyu Yang, Ji Xu, Hao Wu, and Huanjing Yue are with the School of Electrical and Information Engineering, Tianjin
University, Tianjin 300072, China.}
\thanks{
 Kun Li is with the Tianjin Key Laboratory of Cognitive Computing and
Application, College of Intelligence and Computing, Tianjin University,
Tianjin 300350, China.}
\thanks{
 Yu-Kun Lai is with the School of Computer Science and Informatics, Cardiff University, Wales, UK.}
\thanks{
 Jianzhi Lu is with the 3vjia company.}
 \thanks{
Yebin Liu is with the Department of Automation, Tsinghua University, Beijing 10084, China.}
\thanks{
Corresponding author: Kun Li (Email: lik@tju.edu.cn)}}

%
%

\markboth{}%
{Yang \MakeLowercase{\textit{et al.}}: Learning to Reconstruct and Understand Indoor Scenes from Sparse Views}
%



\maketitle

\begin{abstract}
This paper proposes a new method for simultaneous 3D reconstruction and semantic segmentation of indoor scenes. Unlike existing methods that require recording a video using a color camera and/or a depth camera, our method only needs a small number of (\eg, 3-5) color images from uncalibrated sparse views as input, which greatly simplifies data acquisition and extends applicable scenarios. Since different views have limited overlaps,
our method allows a single image as input to discern the depth and semantic information of the scene. The key issue is how to recover relatively accurate depth from single images and reconstruct a 3D scene by fusing very few depth maps. To address this problem, we first design an iterative deep architecture, IterNet, that estimates depth and semantic segmentation alternately, so that they benefit each other. To deal with the little overlap and non-rigid transformation between views, we further propose a joint global and local registration method to reconstruct a 3D scene with semantic information from sparse views. We also make available a new indoor synthetic dataset simultaneously providing photorealistic high-resolution RGB images, accurate depth maps and pixel-level semantic labels for thousands of complex layouts, useful for training and evaluation. Experimental results on public datasets and our dataset demonstrate that our method achieves more accurate depth estimation, smaller semantic segmentation errors and better 3D reconstruction results, compared with state-of-the-art methods.
\end{abstract}



%
\IEEEpeerreviewmaketitle

\section{Introduction}\label{sec:introduction}
With the increasing demand for indoor navigation, home/office design, and augmented reality, indoor 3D reconstruction and understanding have become active topics in computer vision and graphics. Existing reconstruction methods can be broadly categorized into two groups. The first group scans indoor scenes with an integrated depth camera based on either time-of-flight (ToF) or structured light sensing that offers dense measurements of depth. The pioneering KinectFusion~\protect\cite{newcombe2011kinectfusion} presents a detailed workflow using Kinect for indoor reconstruction. It was more recently extended by ElasticFusion~\protect\cite{whelan2016elasticfusion} and BundleFusion~\protect\cite{dai2017bundlefusion} which achieve state-of-the-art results in real-time 3D reconstruction. Despite that it is relatively simple to acquire depth, the depth captured by such methods contains much noise and missing data, and is limited to a small range of distances.
Color cameras do not suffer from these issues, are still far more available (\eg on mobile phones) and have a smaller form factor than depth cameras.
It is therefore interesting to study 3D scene reconstruction using a color camera, which however is challenging due to lack of depth information.
Simultaneous localization and mapping (SLAM)~\protect\cite{SLAMxu} and structure from motion (SFM)~\protect\cite{SFM_IP} are two popular approaches to achieve feature-based point cloud 3D reconstruction on-line and off-line, respectively. However, these feature-based methods require rich textures in the scene, and are therefore difficult to obtain dense point clouds.
All the above methods require consecutive frame tracking or dense view capturing.

\begin{figure}
	\centering
		\includegraphics[width=1.0\linewidth]{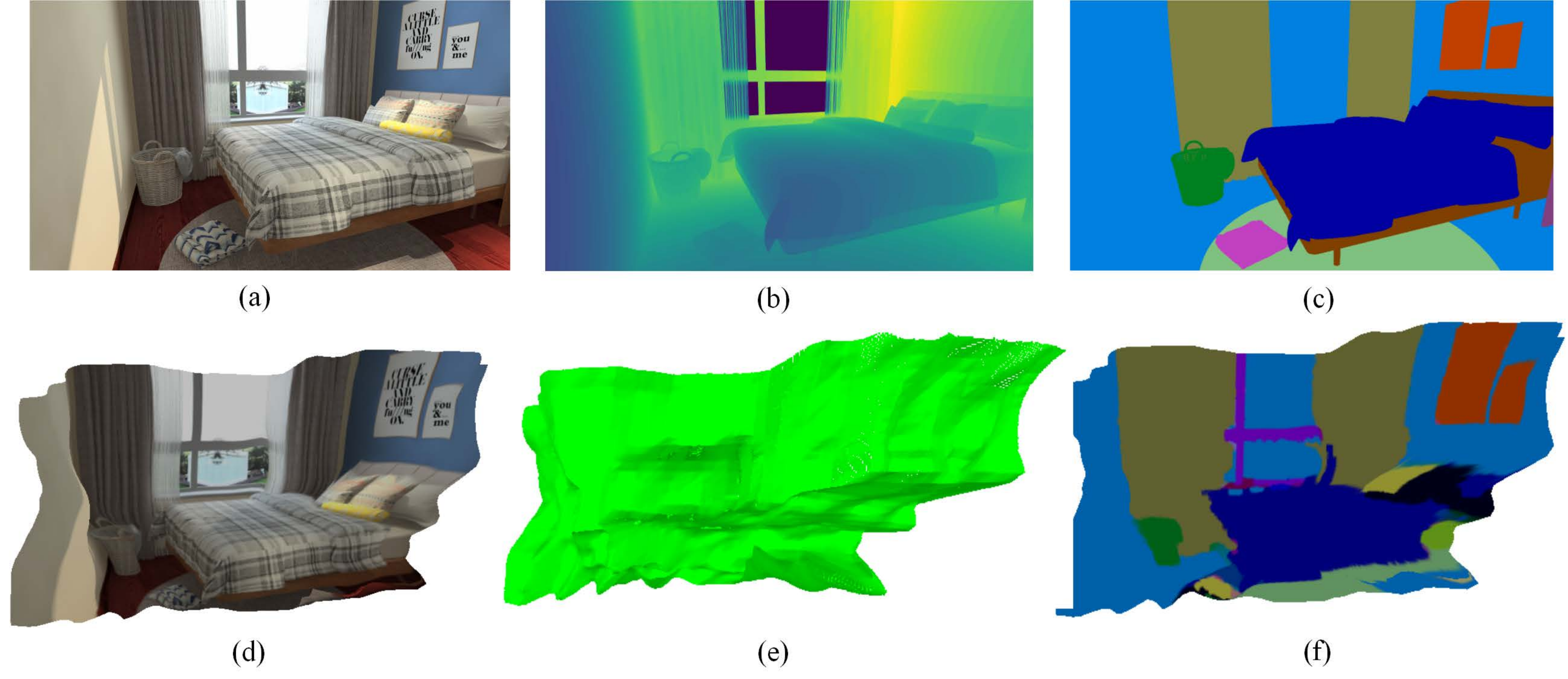}
	\caption{An example of our IterNet RGB-D dataset and the reconstructed 3D model by our IterNet: (a) Input RGB image, (b) Ground-truth depth map, (c) Ground-truth semantic segmentation, (d)-(f) Reconstructed 3D model using our estimated depth map and semantic labels. This example is part of testing data.}
	\label{fig:example}
\vspace{-0.3cm}
\end{figure}

In this paper, we propose a new indoor-scene 3D reconstruction and semantic segmentation method using color images captured from several uncalibrated sparse views. The first challenge is the difficulty in dense reconstruction from sparse views with little overlap, which is practically  degenerated into monocular depth estimation. The second challenge, hence, is non-rigid transformation between views brought in by the inaccurate depth estimated from single color images. To address these problems, we design \emph{IterNet}, an iteratively optimized deep framework for simultaneous depth map recovery and semantic segmentation for each view, where the two tasks help improve each other. To estimate non-rigid transformations between sparse views, we further develop a joint global and local alignment method to fuse estimated depths with the help of semantic information, which integrates geometry, photometry and semantic information in the coarse-to-fine manner.

Depth recovery and semantic segmentation from images are ill-posed and it is essential to learn from high-quality training data.
For indoor scene understanding, a number of datasets have been made publicly available. Real-world datasets, such as NYUv2 \protect\cite{Silberman:ECCV12}, SUN RGB-D \protect\cite{song2015sun} and ScanNet \protect\cite{dai2017scannet}, need a lot of manual labor to annotate the labels and contain unavoidable noise in depths assumed as ground-truths, while synthetic datasets~\protect\cite{song2016ssc,handa2016scenenet} are difficult to generate photorealistic RGB images and usually have limited layouts and image resolution. To our best knowledge, no existing datasets can provide photorealistic RGB images, accurate depth maps, pixel-level semantic labels, and thousands of complex layouts at the same time.
To address this, we build IterNet RGB-D dataset with these features.

Experimental results on both public datasets and our dataset demonstrate that our method outperforms state-of-the-art methods on depth estimation, semantic segmentation, and multi-view reconstruction. Figure~\ref{fig:example} gives an example of our IterNet RGB-D dataset and the reconstructed 3D model with estimated semantics using our IterNet.
We will make the code and the dataset available online for research purposes.

In summary, our work is an integrated work that includes 1) an unprecedented indoor synthetic dataset simultaneously providing photorealistic high-resolution RGB images, accurate depth maps and pixel-level semantic labels for thousands of complex layouts, 2) a depth estimation method from a single color image, 3) a semantic segmentation method from a single view, and 4) a multi-view reconstruction method for sparse views. Each component of our method has novelty and is proved by experiments on public datasets and our dataset. They jointly solve a challenging problem of 3D reconstruction and understanding from sparse views.
Our main contributions are:
\begin{itemize}
\item We provide IterNet RGB-D dataset including photorealistic high-resolution RGB images, accurate depth maps, and pixel-level semantic labels for thousands of complex layouts, useful for training and evaluation.
\item We solve a challenging problem, namely reconstructing and understanding indoor 3D scenes using only color images captured from several uncalibrated sparse views. It is applicable to more scenarios than previous approaches that rely on  texture and/or geometries of dense views, \eg, reconstructing and understanding a room using several photos captured by different users.
\item We design a novel iterative joint optimization method for depth estimation and semantic segmentation for a given input color image, where the two tasks help improve each other.  This architecture is not restricted to these tasks we address here and can also be extended to other related tasks such as object/part parsing.
\item We propose a joint global and local registration method to fuse different sparse perspectives. This coarse-to-fine alignment is robust to the sparsity of views and the errors of monocular depth estimation.
\end{itemize}

\begin{table*}
	\caption{Comparison between various indoor datasets. IterNet RGB-D is our proposed dataset. $\times$: not included, \checkmark : included, -: relevant information not available.}
\renewcommand{\arraystretch}{1.3}
\footnotesize
	\begin{center}
		\begin{tabular}{|c|c|c|c|c|c|c|c|c|}
			\hline
\textbf{Dataset} &  \tabincell{c}{\textbf{NYUv2}\\ \protect\protect\cite{Silberman:ECCV12}} & \tabincell{c}{\textbf{SUN}\\\textbf{RGB-D} \protect\protect\cite{song2015sun}} & \tabincell{c}{\textbf{Building}\\\textbf{Parser} \protect\protect\cite{armeni2017joint}} & \tabincell{c}{\textbf{Matterport}\\\textbf{3D} \protect\protect\cite{chang2017matterport3d}} & \tabincell{c}{\textbf{ScanNet}\\ \protect\protect\cite{dai2017scannet}} & \tabincell{c}{\textbf{SUNCG}\\ \protect\protect\cite{song2016ssc}} & \tabincell{c}{\textbf{SceneNet}\\\textbf{RGB-D} \protect\protect\cite{handa2016scenenet}} & \tabincell{c}{\textbf{IterNet}\\\textbf{RGB-D}}\\
			\hline
			Year & 2012 & 2015 & 2017 & 2017 & 2017 & 2017 & 2016 & 2019\\
			\hline
			Type & Real & Real & Real & Real & Real & Synthetic & Synthetic & Synthetic\\
			\hline
			Images/Scans & 1449 & 10K & 70K  & 194K & 1513 & 130K & 5M & 12,856\\
			\hline
			Layouts & 464 & - & 270 & 90 & 1513 & 45,622 & 57 & 3214\\
			\hline
			Object Classes & 894 & 800 & 13 & 40 & $\ge$ 50 & 84 & 255 & 333\\
			\hline
			RGB & \checkmark & \checkmark & \checkmark & \checkmark & $\times$ & $\times$ & \checkmark & \checkmark \\
			\hline
			Depth & \checkmark & \checkmark & \checkmark & \checkmark & $\times$ & \checkmark & \checkmark & \checkmark\\
			\hline
			Semantic Label & \checkmark & \checkmark & \checkmark & \checkmark & \checkmark & \checkmark & \checkmark & \checkmark\\
			\hline
			RGB Texturing & Real & Real & Real & Real & Real & Not Photorealistic & Photorealistic & Photorealistic\\
			\hline
			\tabincell{c}{Image\\Resolution} & 640$\times$480 & 640$\times$480 & 1080$\times$1080 & 1280$\times$1024 & 640$\times$480 & 640$\times$480 & 320$\times$240 & \tabincell{c}{1280$\times$960;\\1280$\times$720}\\
			\hline
		\end{tabular}
	\end{center}
	\label{table:dataset}
\end{table*}

\section{Related work} \label{sec:related work}

\textbf{Indoor datasets.}  Naseer \etal \protect\cite{naseer2018indoor} gave a comprehensive overview of indoor scene understanding in 2.5/3D. The first dataset is NYU-Depth with two versions introduced by Silberman \etal \protect\cite{Silberman:ECCV12} using Microsoft Kinect. SUN RGB-D dataset \protect\cite{song2015sun} captured by four different RGB-D sensors contains 10,335 indoor images with dense annotations. Armeni \etal \protect\cite{armeni2017joint} provided Building Parser dataset with instance level semantic and geometric annotations. Matterport3D \protect\cite{chang2017matterport3d} contains 10,800 panoramic images covering $360^{\circ}$ viewpoints captured by a Matterport camera. ScanNet \protect\cite{dai2017scannet} is a 3D reconstruction dataset with 2.5 million frames obtained from 1,513 scans. These real-world datasets usually have some noise and missing areas in depth maps and need a lot of manual effort to annotate the labels. Hence, synthetic datasets are proposed for easy generation and accurate ground-truth. SUNCG \protect\cite{song2016ssc} is a densely annotated large-scale indoor dataset, but the rendered RGB images are not photorealistic and RGB-D videos are not available. SceneNet RGB-D \protect\cite{handa2016scenenet} provides pixel-level annotations and photorealistic RGB images, but the number of layouts is limited. Table~\ref{table:dataset} compares various publicly available 2.5/3D indoor datasets with our IterNet RGB-D dataset. Our dataset provides a total of 12,856 photorealistic images for thousands of layouts, and has a higher image resolution: $1280\times960$ and $1280\times720$,
covering more indoor scenes. Moreover, our dataset provides absolute depth maps and pixel-level semantic segmentation that are more precise and accurate. Compared with other datasets, the indoor scenes covered by our dataset are more general and more complex.

\textbf{Monocular Depth Estimation.} In computer vision, monocular depth estimation has been a long-standing topic in the last decades. Previous approaches mainly focused on hand-crafted features \protect\cite{hoiem2005automatic}, defocused features\protect\cite{DepthDefocus}, statistical priors\protect\cite{DepthNSS} or graphical models \protect\cite{liu2014discrete}. With the development of deep learning, more recent approaches are based on Convolutional Neural Networks (CNNs). For instance, Eigen \etal~\protect\cite{eigen2014depth} proposed a multi-scale CNN for depth estimation and demonstrated the effectiveness of the CNN-based method with promising results. Considering the correlation between tasks, Wang \etal \protect\cite{wang2015towards} introduced a CNN for joint depth estimation and semantic segmentation. Xu \etal \protect\cite{xu2018pad} proposed a multi-task approach for depth estimation via cross-modal interactions to refine the task. Recently, the attention mechanism has become popular, and Xu \etal \protect\cite{xu2018structured} proposed a structured attention mechanism to fuse the features of different scales. The most similar work to ours is \protect\cite{xu2017multi} where a continuous Conditional Random Field (CRF) is used to combine multi-scale features. Our approach develops from a similar intuition but further integrates semantic information in an iterative way.

\textbf{Semantic Segmentation.} Semantic segmentation is an extension of image classification. Instead of classifying an image as a whole, semantic segmentation assigns per-pixel predictions of object categories for the given image. It is challenging due to randomness of object distribution, poor illumination, and occlusion. Deng \etal \protect\cite{Segmentation2016} proposed a robust information theoretic (RIT) model to reduce the uncertainties, \ie, missing and noisy labels, by learning a transformation function and a discriminative classifier that maximize the mutual information of data and their labels in the latent space. Alterative approaches are typically based on CNNs. Long \etal \protect\cite{long2015fully} proposed a Fully Convolutional Network (FCN), a popular CNN architecture for dense predictions without any fully connected layers. Almost all the subsequent approaches on semantic segmentation adopted this paradigm. With the development of depth sensors and the release of RGB-D datasets, some methods attempted to use depth information for better segmentation, no longer limited to a single RGB image. Li \etal \protect\cite{li2016lstm} constructed HHA images \protect\cite{gupta2014learning} for the depth channel through geometric encoding before feeding them to the network and used Long Short-Term Memory (LSTM) to fuse two different features. Ma \etal \protect\cite{Ma2017Multi} predicted semantic segmentation from RGB-D sequences, but it is inapplicable to sparse views. Our method exploits depth information to help improve semantic segmentation, but the depth is estimated from the input color image instead of directly captured by a dedicated depth sensor. We propose an iterative method for joint estimation of the depth and semantic segmentation, which benefit each other.

\textbf{Indoor Scene 3D Reconstruction.} Indoor Scene 3D Reconstruction from a color video or multi-view color images is a challenging and active topic. Given a color video, most structure from motion (SFM) methods \protect\cite{snavely2006photo} recovered the 3D structure by estimating the motion of the cameras corresponding to the frames. However, it is difficult for these methods to obtain dense and accurate reconstruction. Given multi-view color images with calibrated camera parameters, multi-view stereo (MVS) methods \protect\cite{liu2010point} can achieve more accurate 3D reconstruction. But they require adjacent views to have sufficient overlap and cannot work well with sparse views. COLMAP~\protect\cite{schoenberger2016sfm,schoenberger2016mvs} provides a pipeline containing both SFM and MVS with graphical and command-line interfaces. When the views of images are very sparse, the depth of each image can be estimated and fused together using iterative closest point (ICP) like registration methods \protect\cite{Geiger2012CVPR}. However, it is difficult to achieve accurate depth estimation from individual color images which increases the difficulties of ICP fusion. Saxena \etal \protect\cite{Saxena20073} proposed a novel method for 3D reconstruction from sparse views, but it only worked well for building-like outdoor scenes and cannot generate semantics. Learning-based methods, \eg, MVSNet \protect\cite{Yao2018MVSNet} and DeepMVS \protect\cite{DeepMVS}, output the depth of a specific frame based on a color multi-view sequence, but they cannot deal with sparse views. In this paper, we design \emph{IterNet} to estimate a more accurate depth map with the help of semantic segmentation, and propose a joint global and local registration method to better achieve indoor scene 3D reconstruction from sparse views.

\section{Proposed Method}
\begin{figure*}
	\centering
		\includegraphics[width=0.9\linewidth]{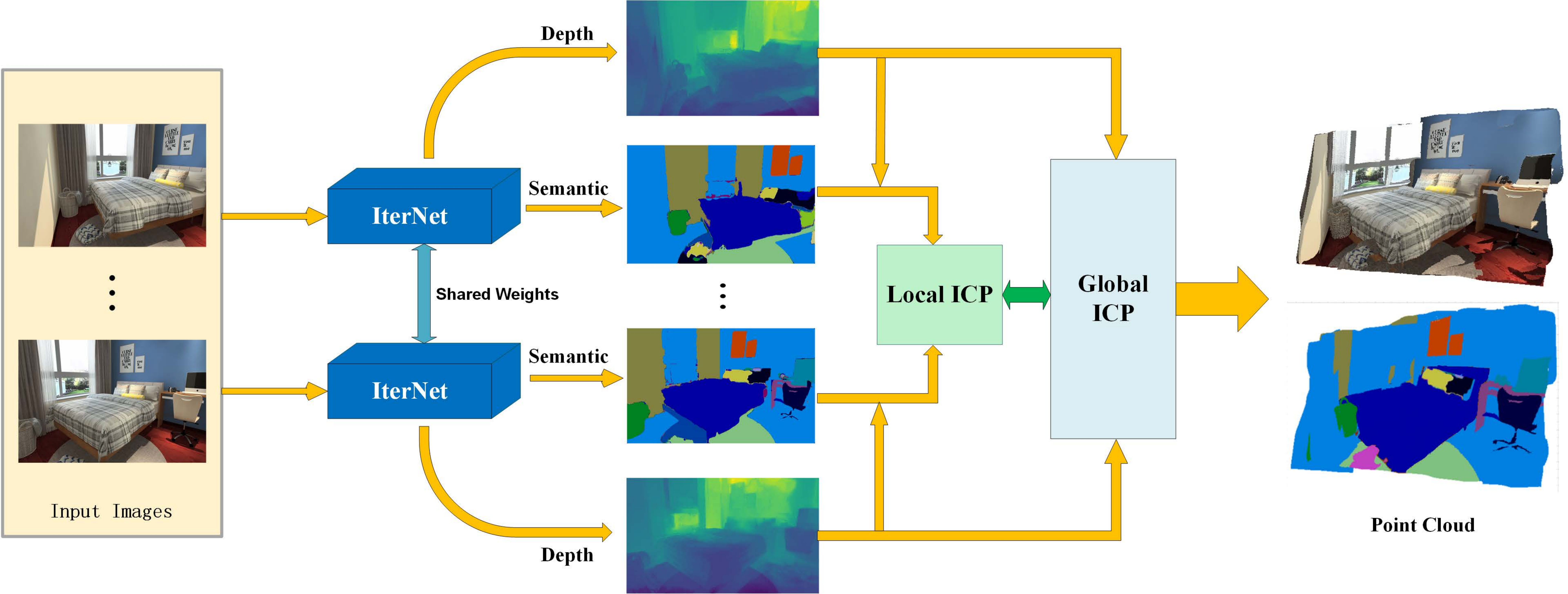}
	\caption{Illustration of the proposed method for indoor 3D reconstruction and understanding. The blue Module refers to our IterNet for iterative joint depth estimation and semantic segmentation (Section \ref{jointMethod}). With the help of semantic segmentation, we use our proposed joint global and local registration method to reconstruct a 3D scene with semantic information from sparse views (Section \ref{localICP}).}
	\label{fig:framework}
\end{figure*}

  In this section, we first introduce our IterNet RGB-D dataset in Section \ref{dataset}, and then describe the technical details of IterNet for iterative joint depth estimation and semantic segmentation in Section \ref{jointMethod}. The joint global and local multi-view reconstruction method is presented in Section \ref{localICP}. Figure~\ref{fig:framework} illustrates the workflow of our method.

\subsection{Dataset}
\label{dataset}
\begin{figure}[t]
	\centering
		\includegraphics[width=1.0\linewidth]{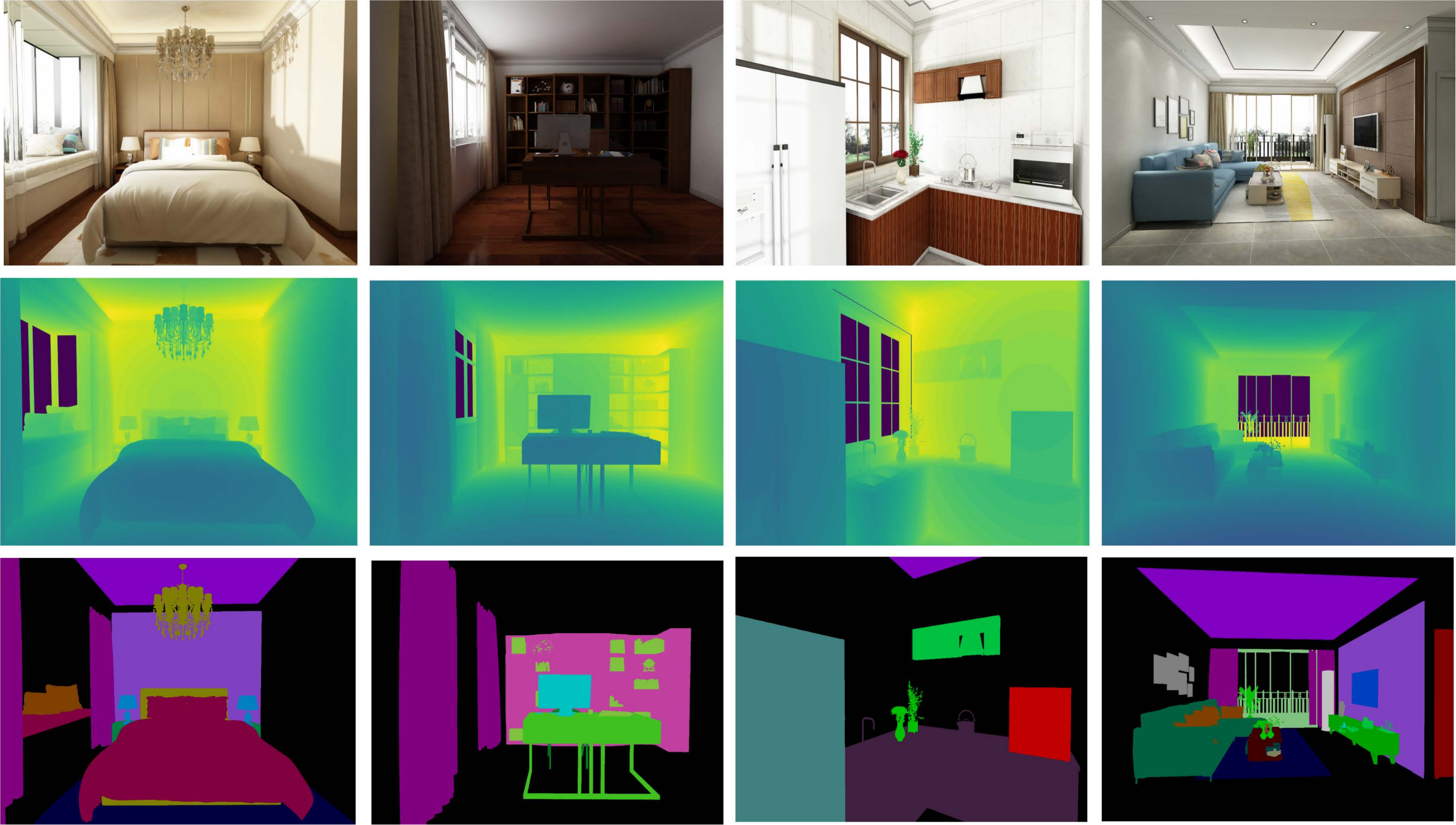}
	\caption{Some examples of different scenarios in our dataset. From top to bottom: color images, ground-truth depth maps, and ground-truth semantic segmentations.}
	\label{fig:dataSample1}
\vspace{-0.4cm}
\end{figure}

\begin{figure*}[!ht]
 	\centering
 	\includegraphics[width=0.9\linewidth]{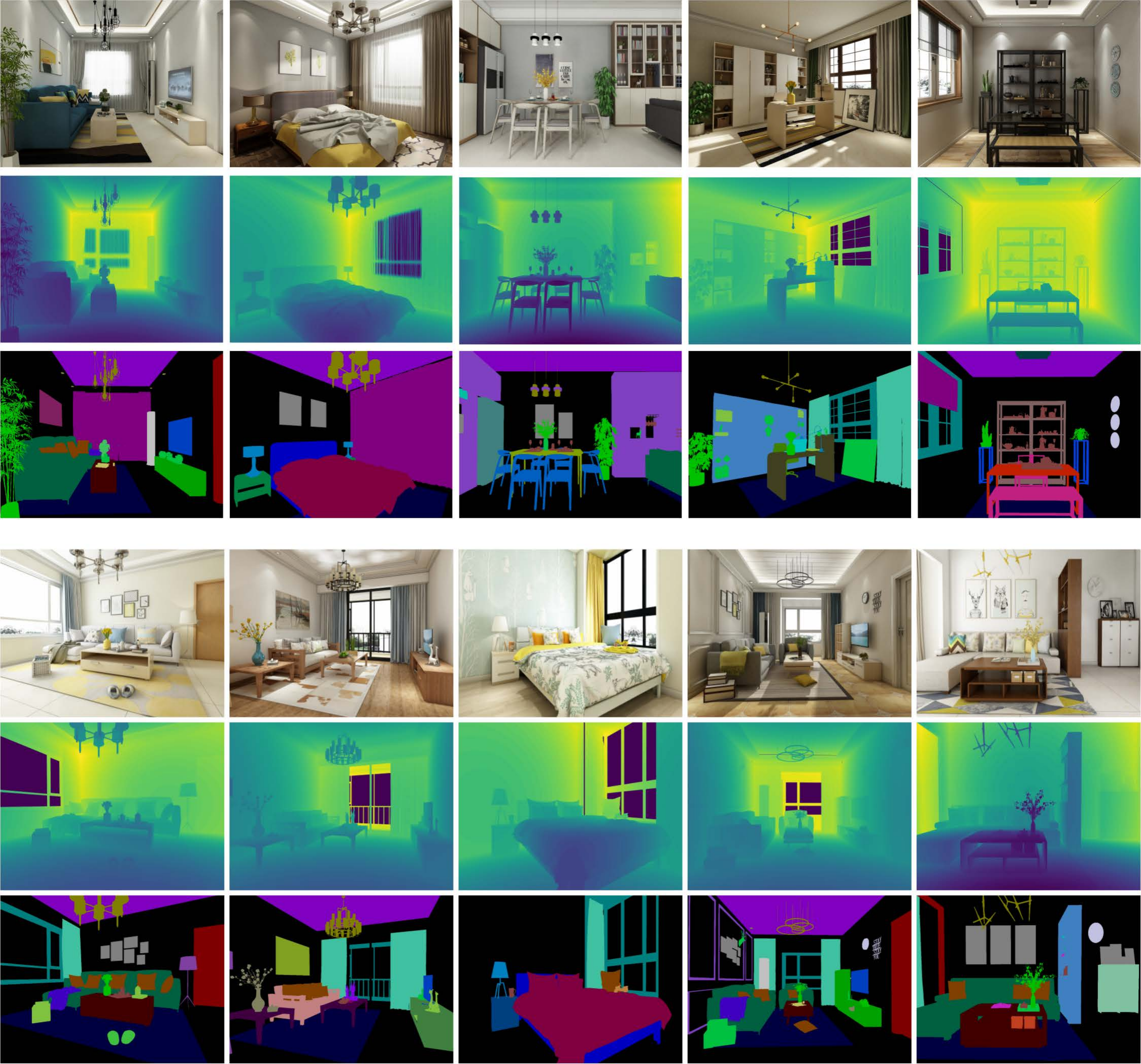}
 	\caption{More examples of different scenarios in our dataset with color images, depth maps, and semantic labels. Our dataset contains more complex indoor layouts, richer textures, colorful and realistic lightings, and higher resolution images.}
 	\label{fig:dataSample}
  \end{figure*}

Different from the production of other synthetic datasets \protect\cite{handa2016scenenet,song2016ssc}, our dataset is generated by a third-party platform which includes various real-life house styles, real prototype rooms designed by professional designers, and detailed model materials. We also implement high-quality photorealistic rendering. Compared to traditional rendering, we adopt the method of image splitting and recombination to achieve distributed rendering. To accelerate the rendering speed, we utilize the computing power of multiple servers with CPUs, thus multiplying the rendering speed. The average rendering time of a $1280\times960$ image is about 90 seconds. Our rendering is realized on a cluster of 32 servers, each consisting of a CPU with 32 cores and 64 threads. For rendering 12,856 images, it takes about 321 hours. In terms of rendering quality, in addition to considering the direct illumination of the light source in the scene, the illumination reflected by other objects, known as Global Illumination (GI), is also taken into consideration. There are many ways to achieve GI. In order to render better results, we adopt the Brute Force (BF) algorithm \protect\cite{cook1984distributed} based on path tracking. The number of samples per pixel is up to 512 and varies for different scenes. The noise level is controlled below 0.05. A lower noise level yields better rendering quality, but requires longer rendering time. In order to obtain better results and minimize the rendering time, rendered images are denoised using a wavelet-based denoising method \protect\cite{donoho1994ideal}. Figure \ref{fig:dataSample1} shows some examples of different scenarios in our dataset. Our dataset provides photorealistic high-resolution RGB images, accurate depth maps and pixel-level semantic labels for thousands of layouts, useful for training and evaluation. Figure \ref{fig:dataSample} shows more scenarios in our dataset. It can be seen that our dataset contains more complex indoor layouts, richer textures, colorful and realistic lightings, and higher resolution images, which are more photorealistic and closer to real-world images than existing synthetic datasets. Our dataset will be available online.

\subsection{IterNet: Iterative CNN for Joint Depth Estimation and Semantic Segmentation}
\label{jointMethod}
\begin{figure*}
	\centering
		\includegraphics[width=0.95\linewidth]{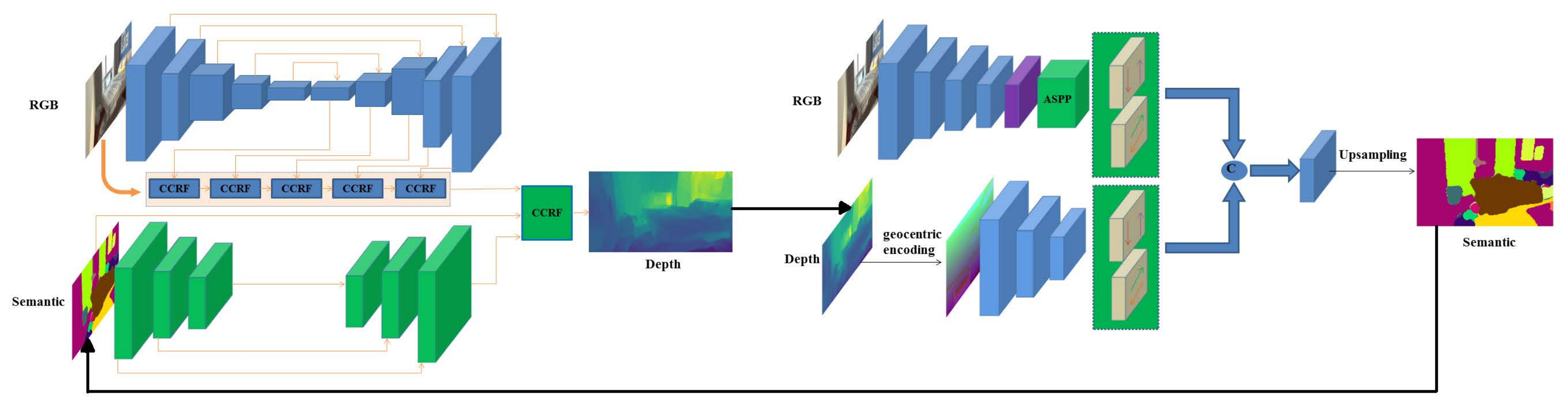}
		\caption{Overview of the proposed IterNet architecture. The CCRF blocks in the depth estimation sub-network fuse the features at different scales and combine the semantic features. In the semantic segmentation sub-network, the purple block represents atrous convolution which reduces the size of image while increasing the receptive field. The ASPP block indicates atrous spatial pyramid pooling which is made by four different dilated convolutions for resampling in our implementation. }
	\label{fig:joint}
\vspace{-0.3cm}
\end{figure*}

\textbf{Network Architecture.} The proposed IterNet is a multi-task deep CNN mainly consisting of two parts: the depth estimation sub-network and the semantic segmentation sub-network, as shown in Figure~\ref{fig:joint}.

In the design of the depth estimation sub-network, we refer to a monocular depth estimation method \protect\cite{xu2017multi} using a continuous conditional random field (CCRF) to combine multi-scale features. Different from \protect\cite{xu2017multi}, we add a semantic branch built upon an encoder-decoder structure to extract semantic features and further use a CCRF to integrate the multi-scale RGB features and the semantic features which can better make use of boundary constraints in semantic segmentation. The RGB branch consists of a front-end base network and a refinement network combined with several CCRF modules. Together with semantic information, the output of the RGB branch is fed into a CCRF module to generate the estimation of depth which is used as the input of the semantic segmentation sub-network.

In the semantic segmentation sub-network, we use the Long Short-Term Memorized Context Fusion (LSTM-CF) \protect\cite{li2016lstm} Model with different fusion scheme for the RGB-D features, which is capable of fusing contextual information from multiple sources (\ie~photometric and depth channels). Instead of the original serial vertical and horizontal context layers, we adopt a parallel context layer and a direct fusion scheme to better play the role of depth. We also add an Atrous Spatial Pyramid Pooling (ASPP) \protect\cite{chen2018deeplab} as a multi-scale feature extractor. Unlike an encoder-decoder network extracting different intermediate layers to obtain multi-scale features, ASPP employs multiple parallel filters with different sampling rates. For depth information, rather than directly feeding a depth image into the network, we first encode it into an HHA image \protect\cite{gupta2014learning} using geocentric encoding and then input it into the network.


\textbf{Training and Testing.}
Given datasets of  \emph{RGB-Depth-Semantic} triplets, our aim is to train the designed network for joint depth and semantic estimation. The depth estimation sub-network and semantic segmentation sub-network are designed to interact with each other to boost the performance. Instead of jointly training the two sub-networks, we train the depth estimation and semantic segmentation sub-networks sequentially for flexible boosting. Taking the depth estimation sub-network as an example, we train the upper branch and the lower branch with \emph{RGB-Depth} pairs and \emph{Sematic-Depth} pairs, respectively. The depth estimation sub-network is then fine-tuned with the  \emph{RGB-Depth-Semantic} triplets. The semantic segmentation sub-network is trained in a similar way.

At the test stage, since each sub-network expects the output of the other sub-network as part of input, we use the following strategy. We need an initialized semantic segmentation or depth estimation which can be easily obtained by disabling one of the branches in the original network structure. For example, if we want to obtain an initial depth estimation for semantic segmentation, we disable the semantic segmentation branch in the depth estimation sub-network and then extract features from RGB branches as an initial depth. We then alternately run the two sub-networks, with the output of one sub-network used as input for the other sub-network. The additional depth information helps improve semantic segmentation, and the semantic segmentation in turn contributes to improved depth estimation. In practice, we find that there is no significant improvement after 3 iterations, which shows quick convergence.

\textbf{Implementation Details.}
The proposed approach is implemented on the Caffe framework \protect\cite{jia2014caffe} and runs on a computer with an Nvidia GTX 1080ti graphics card (11GB). For depth estimation sub-network, the learning rate is initialized at $10^{-11}$ and decreases by $10\%$ for every 30 epochs. The batch size is set to 16. The momentum and the weight decay are set to 0.9 and 0.0005, respectively. The semantic segmentation sub-network follows the same training rules, but the initial learning rate is set to $10^{-4}$. The parameters of batch size, momentum and weight decay are set to 8, 0.9 and 0.005, respectively. The learning rate decreases by $10\%$ for every 20 epochs. When the pretraining of each branch is finished, we fine-tune the sub-networks, and the initial learning rates are set to $10^{-12}$ and $10^{-5}$ for depth estimation and semantic segmentation, respectively. The batch size, momentum and weight decay remain the same as the pretraining.

\subsection{Joint Global and Local Reconstruction} \label{localICP}

After obtaining the depth and the semantic segmentation for the image of each view, we reconstruct the whole 3D scene by fusing the depths of different views. The straightforward way is to use the ICP algorithm to align the point clouds transformed from the depths of different perspectives. However, it is difficult to achieve satisfactory alignment. First, the depths are obtained by a monocular depth estimation network, not captured by Kinect or other depth cameras, containing some non-statistical errors. It is therefore insufficient to align two depth point clouds with just one rigid transformation. Second, for sparse perspectives, the overlap between two adjacent views is limited which is difficult to handle by standard ICP algorithms. Hence, we propose a new joint global and local registration method by exploiting photometric and semantic information to improve reconstruction quality.

Before fusion, we filter the messy points based on the plane constraint similar to \protect\cite{bodis2015superpixel}. Let $\mathcal{X} \triangleq \{X_i\} = \{ ( C_i, D_i, S_i  ) \}_{i=1}^N$ be the sparse view set, where $N$ is the total number of views for reconstruction. After depth estimation and semantic segmentation, each view now contains three components: color $C_i$, depth $D_i$ and segmentation $S_i$. We align all the depth point clouds in sequence with the previous
registration result used as the next target model. Each alignment has two stages, namely global alignment and local alignment.

\textbf{Global alignment.}
Taking the point cloud generated using the previous $i-1$ views as the target, our goal for global alignment is to find an optimal global rigid transformation $\mathcal{T}_i$ for view $i$, which is composed of two parts: rotation $R_i$ and translation $t_i$. Specifically, we first convert the depth map $D_i$ into a point cloud $\mathcal{P}_i=\{p_1,p_2,...,p_{n_i}\}, i \in \{1, 2, \dots, N\}$. $\mathcal{P}_i$ is a point set for the \emph{i}-th view, and $n_i$ represents the total number of points in the view.
We take a global ICP-type framework alternating two steps, until convergence. The transformation is initialized by a $4\times4$ identity matrix. Assuming the target point cloud is $\mathcal{P}_j$ containing all the fused points from the previous views, the first step finds for each point $p_k \in \mathcal{P}_i$ its corresponding point $p'_{\tilde{k}} \in \mathcal{P}_j$ if possible, and the second step updates the transformation $\mathcal{T}_i$ such that when applied to $\mathcal{P}_i$ the point cloud is aligned with $\mathcal{P}_j$.


In the first step, we exploit the additional photometric and semantic information. We lift each point $p_k \in \mathcal{P}_i$ from 3D to a point in a 7-dimensional (7D) space, $\hat{p}_k=(x_k,y_k,z_k,r_k,g_k,b_k,s_k)$, including its 3D position $(x_k,y_k,z_k)$, RGB color $(r_k,g_k,b_k)$ and semantic label $s_k$. Similarly, the point $p'_{\tilde{k}} \in \mathcal{P}_j$ is lifted to a 7D point $\hat{p}'_{\tilde{k}}$.
Our global registration method for aligning $\mathcal{P}_i$ and $\mathcal{P}_j$ first finds the corresponding point $p'_{\tilde{k}} \in \mathcal{P}_j$ for each point $p_k$ in $\mathcal{P}_i$ by the following optimization:
\begin{equation} \label{equation1}
\begin{split}
\tilde{k} = \argmin_{v \in \{1,2,...,n_j\}} ( {\| p_k(x,y,z)-p'_v(x,y,z) \|}^2 \\ +  w_1 {\|p_k(r,g,b)-p_v^{'}(r,g,b)\|}^2 \\
+ w_2 {\| p_k(s)-p_v^{'}(s) \|}^2),
\end{split}
\end{equation}
where $w_1$ and $w_2$ are weights to balance the importance of geometric, photometric and semantic information. They are set to be $w_1=0.1$ and $w_2=10$ in our experiments.

Due to limited overlap, not all the points in $\mathcal{P}_i$ have their corresponding points in $\mathcal{P}_j$. We reject $p'_{\tilde{k}}$ if the matching error is larger than a threshold. In our implementation, this threshold is set to 5cm, and correspondences with higher distances are ignored.
Let $\mathcal{C}_{i,j} = \{ p_k, p'_{\tilde{k}} \}$ be the set of retained correspondences.
In the second step, since photometric and semantic matching errors are independent of rigid transformations,
we use a standard ICP algorithm \protect\cite{Geiger2012CVPR} to find the transformation between the two point clouds:
\begin{equation} \label{equation2}
\begin{split}
(R_i, t_i) = \argmin_{R, t} \frac{1}{2}\sum_{(p_k, p'_{\tilde{k}}) \in \mathcal{C}_{i,j})}{\lVert p'_{\tilde{k}}-Rp_k-t \rVert}_2^2.
\end{split}
\end{equation}

\begin{figure*}[!t]
	\centering
		\includegraphics[width=0.8\linewidth]{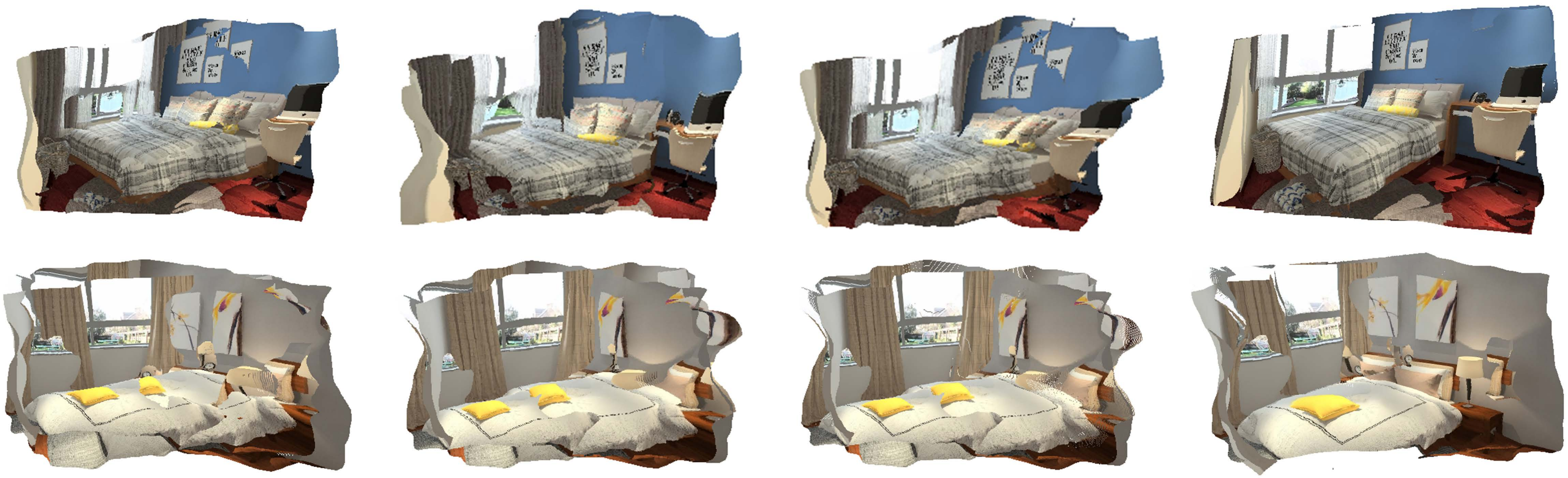}
	\caption{Comparison of different alignment methods: From left to right are results of standard ICP algorithm \protect\protect\cite{Geiger2012CVPR}, 4PCS~\protect\protect\cite{amo_fpcs_sig_08}, global alignment using the estimated depth without the help of semantic branch, and our joint global and local alignment method.}
	\label{fig:localICP}
\vspace{-0.2cm}
\end{figure*}

\textbf{Local alignment.}
Using the 7D global registration method, we achieve coarse alignment which broadly aligns different views, but still cannot cope with the problem of non-statistical errors in monocular depth estimation, as such local deformation is no longer rigid.
To address this problem, we further propose a local registration strategy to refine the previous coarse estimation, similar to coarse-to-fine refinement.
Specifically, we first extract local point sets from the original point cloud according to their semantic labels, and then register each of them using the above method. 
Note that in this case, a subset of points from one view is only matched to subsets of points with the same semantic label. Therefore, when finding the matched point, the semantic difference term in Eq.~(\ref{equation1}) is always zero.
For each local set, once it is aligned, we fuse the registered parts from different views by averaging 3D positions of overlaps to mitigate the influence of noise.
The key for our joint global and local registration method is to use multiple transformations to register sparse views with coarse-to-fine refinement, rather than just one single transformation, which is more robust to the noise and outliers in the monocular depth estimation.

\section{Experimental Results}

\subsection{Ablation Study}
We compare the full model with full model without semantic segmentation and full model without depth estimation in Table \ref{tab:ablation}. It can be seen that our full model has achieved the best performance. Figure \ref{fig:localICP} shows the fusion results of an ICP matching method \protect\cite{Geiger2012CVPR}, 4PCS \protect\cite{amo_fpcs_sig_08}, global alignment using the estimated depth without the help of semantic branch, and our proposed joint global and local registration method.  Some misalignments occur in local areas for standard ICP methods. On the contrary, our method achieves better fusion result in terms of both global structure and local details.

\begin{table}[!h]
\caption{Ablation study on our dataset. F-S: full model without semantic; F-D: full model without depth; F: full model.}
	\renewcommand{\arraystretch}{1.3}
	\scriptsize
\centering
\begin{tabular}{|c|c|c|c|}
  \hline
  Method & F-S & F-D & F  \\
  \hline
  rel (lower is better)  & 0.176 & - & \textbf{0.136} \\
  log10 (lower is better) & 0.088 & - & \textbf{0.062} \\
  rms (lower is better) & 1.012 & - & \textbf{0.507} \\
  \hline
  P-acc.(\%) (higher is better)  & - & 67.35 & \textbf{75.54} \\
  M-acc.(\%) (higher is better) & - & 68.29 & \textbf{74.49} \\
  IoU(\%) (higher is better) & - & 54.21 & \textbf{63.98} \\
  \hline
\end{tabular}
\label{tab:ablation}
\end{table}

\begin{figure}[t]
	\centering
	\includegraphics[width=1.0\linewidth]{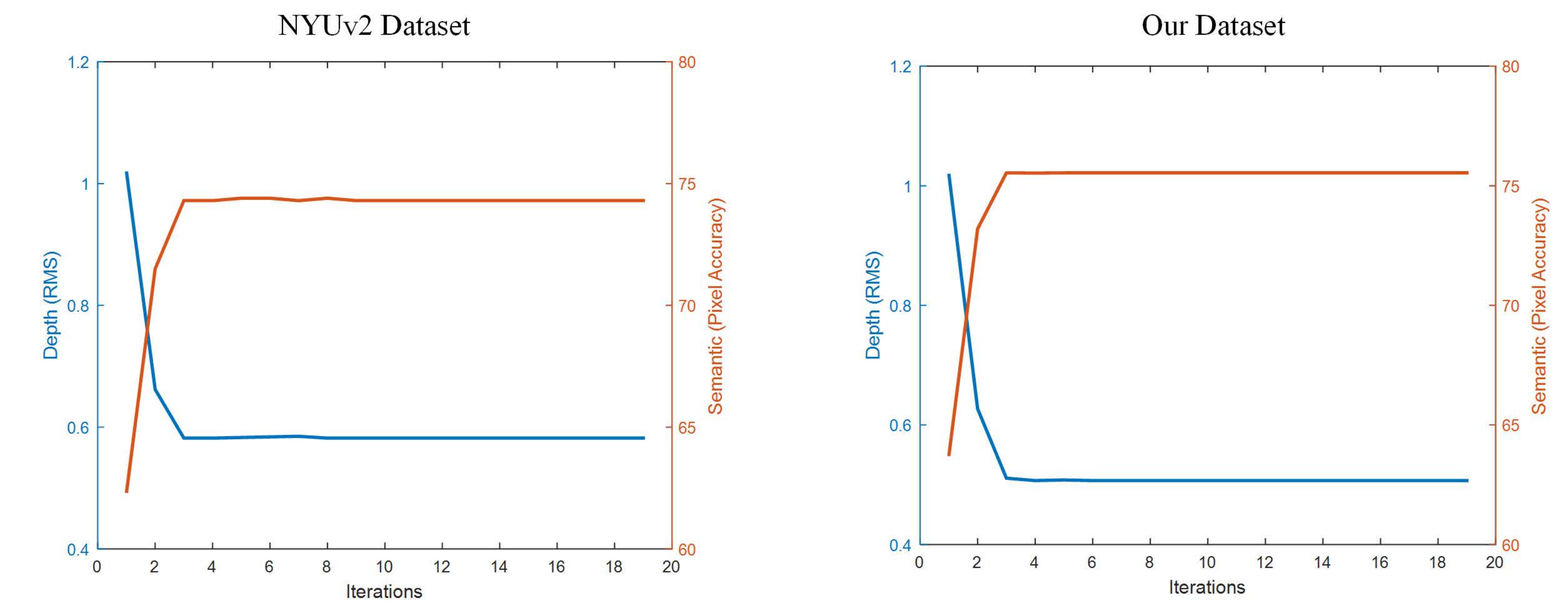}
	\caption{Convergence curves of the proposed IterNet for NYUv2 dataset \protect\cite{Silberman:ECCV12} and our dataset (averaged over all test images in each dataset).}
	\label{fig:iteration}
\end{figure}

Our iterative scheme in IterNet usually converges to promising results after three iterations and is stable for various images. Figure \ref{fig:iteration} shows the decreasing of average RMS (root mean squared) errors of depth estimation over all the test images in iterations and the increasing of average pixel accuracy of semantic segmentation over all the test images in iterations. It can be seen that there is no significant improvement for both depth estimation and semantic segmentation beyond three iterations.

\begin{figure*}[htbp]
	\centering
		\includegraphics[width=0.9\linewidth]{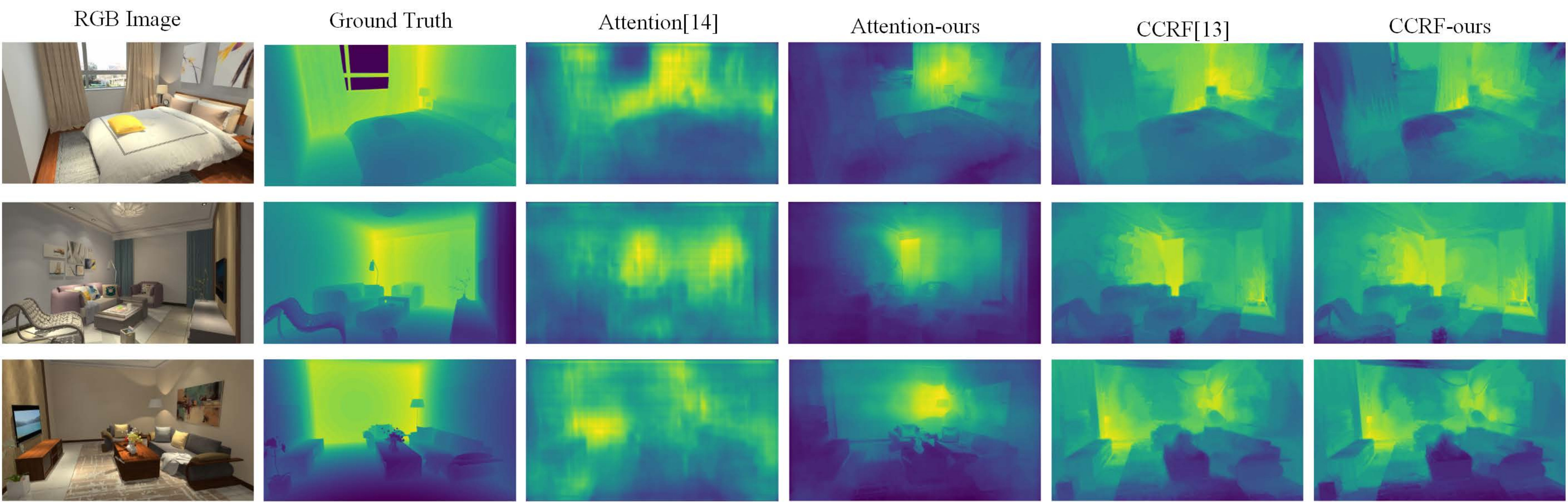}
		\caption{Comparison of depth estimation with two different network architectures. }
	\label{fig:depthBackbone}
\end{figure*}

To study and verify the role of IterNet in depth estimation, we compare two recent backbone architectures including  Structured Attention Guided Convolution Neural Fields \protect\cite{xu2018structured} and CCRF \protect\cite{xu2017multi} which achieve promising performance in depth estimation. Figure \ref{fig:depthBackbone} shows the comparison results on our IterNet RGB-D dataset. We crop high resolution images into small pieces of 426 $\times$ 426 and feed them into the networks. It can be seen that our framework significantly enhances the attention with clear object structures, and refines the CCRF architecture with sharper contours for some objects such as the pillow and the chair.

\begin{figure*}
	\centering
		\includegraphics[width=0.85\linewidth]{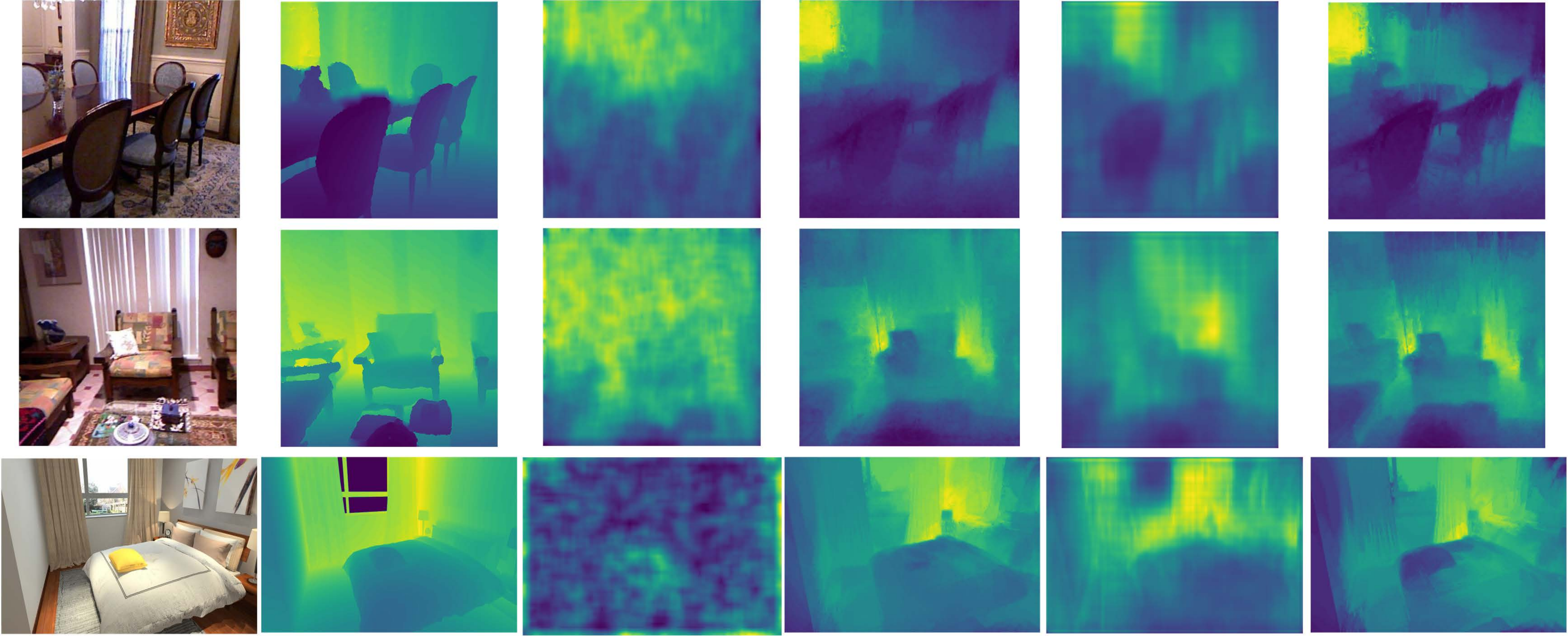}
		\caption{Depth estimation results on NYUv2 dataset (top two rows) and our dataset (bottom row).
		From left to right are the input RGB images, the ground-truths depth and the depth results estimated by Eigen \etal \protect\cite{eigen2014depth}, Xu \etal \protect\cite{xu2017multi}, Xu and Wang \protect\cite{xu2018structured}, and our method.}
	\label{fig:depth_all}
\vspace{-0.4cm}
\end{figure*}

\subsection{Depth Estimation}
\begin{figure*}[htbp]
	\centering
	\includegraphics[width=0.7\linewidth]{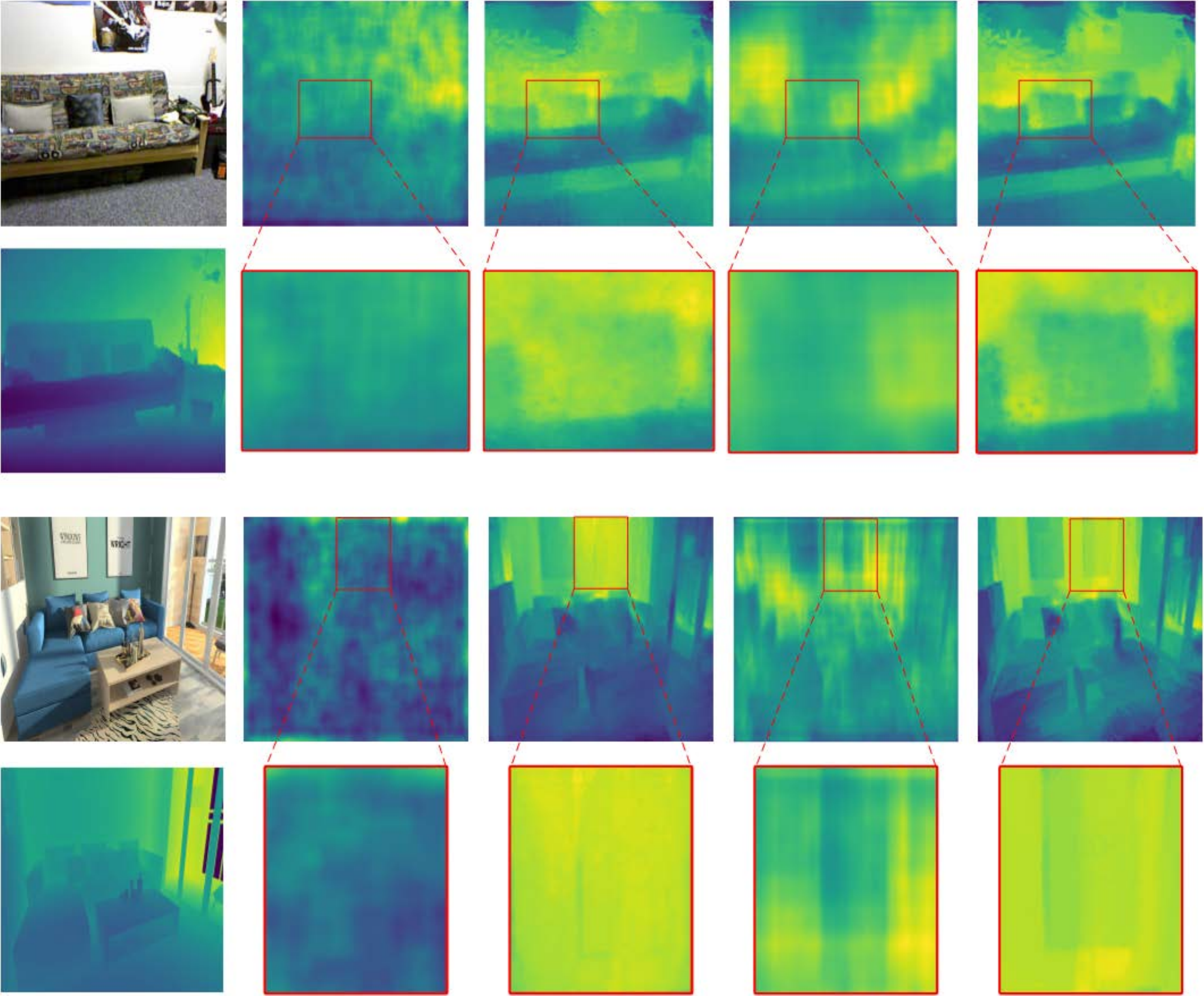}
	\caption{Depth estimation results on NYUv2 dataset \protect\cite{Silberman:ECCV12} and our dataset. From left to right are the input RGB images and the ground-truth depths, the depth results estimated by Eigen \etal \protect\cite{eigen2014depth}, the depth results estimated by Xu \etal \protect\cite{xu2017multi}, the depth results estimated by Xu and Wang \protect\cite{xu2018structured}, and the depth results estimated by our method.}
	\label{fig:depth_nyu}
\end{figure*}

We compare our approach with several state-of-the-art methods on NYUv2 dataset \protect\cite{Silberman:ECCV12} in Table \ref{depthExperiment}. We use 795 images for training and the other 654 images for testing as other methods did. We also use the same raw data as other methods and adopt data augmentation (finally 4770 images for training) to avoid the over-fitting problem. Referring to previous work \protect\cite{eigen2015predicting,eigen2014depth,wang2015towards}, we evaluate the depth estimation results with the following metrics: (1) mean relative error (rel): $\frac{1}{P}\sum_{i}^{}\frac{\left| d_i-d_i^* \right|}{d_i^*}$; (2) root mean squared error (rms): $\sqrt{\frac{1}{P}\sum_{i}^{}(d_i-d_i^*)^2}$; (3) mean log10 error (log10): $\frac{1}{P}\sum_{i}^{}\lVert \log_{10}(d_i)-\log_{10}(d_i^*)\rVert$ and (4) accuracy with threshold $t$: percentage($\%$) of $d_i^*$ subject to max$(\frac{d_i^*}{d_i},\frac{d_i}{d_i^*})=\delta < t$, where $d_i$ and $d_i^*$ denote the predicted depth value and the ground-truth value for pixel $i$. $P$ is the total number of pixels. The results of the compared methods are quoted from their papers. Our method outperforms thirteen competing methods in all metrics, and is comparable to  PAD-Net \protect\cite{xu2018pad} which has a more complex network structure  and requires ground-truth contours and normals as part of labels. We run multiple training trials and consistently achieve the results. We also quantitatively evaluate some methods with their provided code on our IterNet RGB-D dataset. As shown in Table \ref{depthComp}, our method achieves the most accurate depth estimation on all the metrics. Figure \ref{fig:depth_all} gives some visual comparison results on NYUv2 dataset \protect\cite{Silberman:ECCV12} and our dataset. Figure \ref{fig:depth_nyu} gives more qualitative comparison results with enlarged local areas on NYUv2 dataset \protect\cite{Silberman:ECCV12} and our dataset. It can be seen that our method achieves more accurate depth estimation consistent with the quantitative evaluation. Although \protect\cite{xu2017multi} also has good visual results due to promising estimation of relative depths between objects, our method achieves more accurate results both visually and quantitatively.

\begin{figure*}[htbp]
	\centering
		\includegraphics[width=0.8\linewidth]{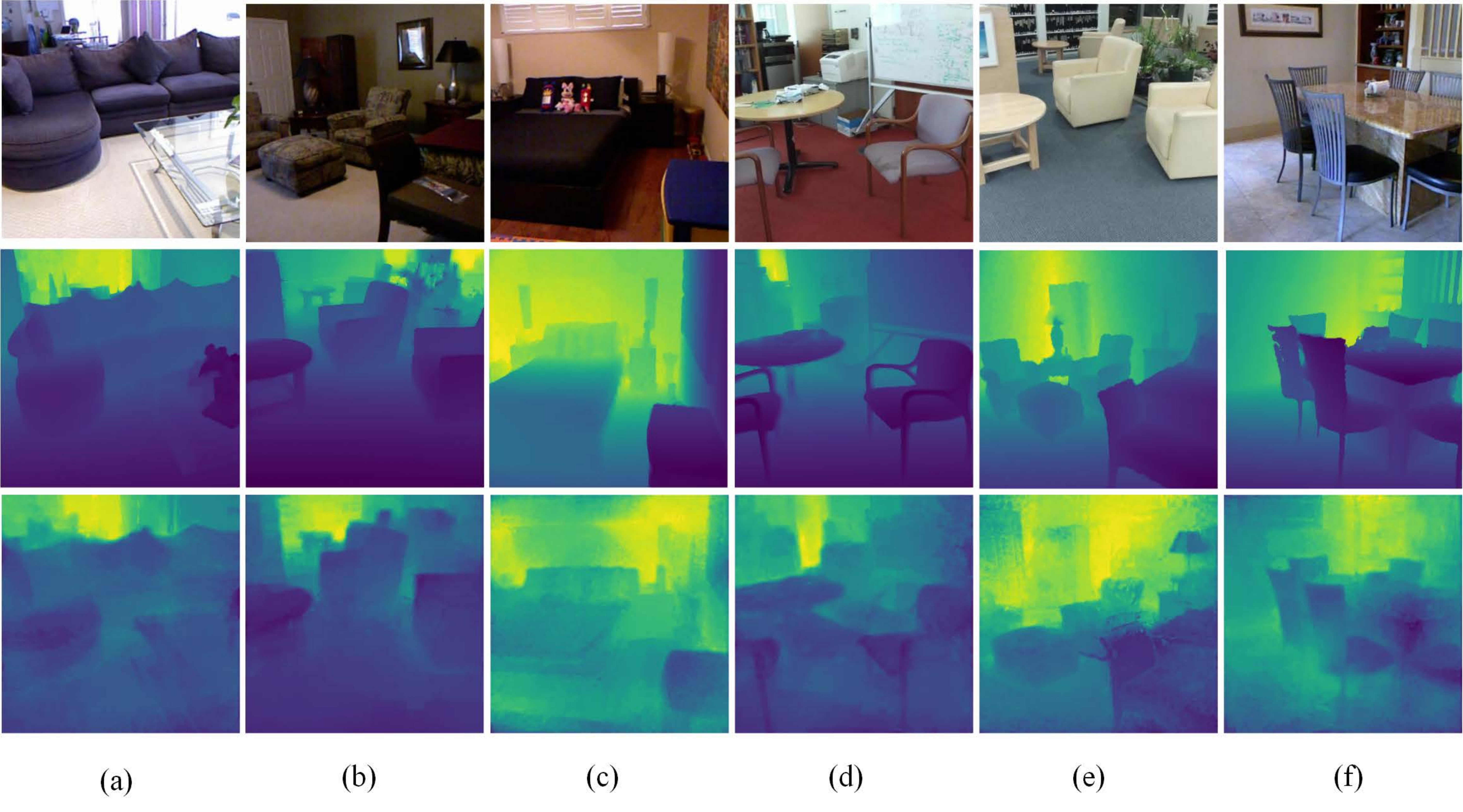}
		\caption{Depth estimation results on NYUv2 dataset \protect\cite{Silberman:ECCV12} (a, b, c) and SUN RGB-D dataset \protect\cite{song2015sun} (d, e, f) using our model trained by our dataset. From top to bottom are the input color images, the ground truths, and our estimated depths.}
	\label{fig:depthGeneral}
\end{figure*}

To evaluate the generalizability of our model trained by our dataset, we show some depth estimation results for real indoor scenes on NYUv2 dataset \protect\cite{Silberman:ECCV12} and SUN RGB-D dataset \protect\cite{song2015sun} without finetuning in Figure \ref{fig:depthGeneral}. It can be seen that our model trained using our dataset has good generalization ability to other datasets.

\begin{table}
\caption{Quantitative evaluation for depth estimation on NYUv2 dataset.}
	\renewcommand{\arraystretch}{1.3}
	\small
	\begin{center}
		\resizebox{1.0\linewidth}{!}{
			\begin{tabular}{|c|c|c|c|c|c|c|c|}
				\hline
				\multicolumn{2}{|c|}{\multirow{2}{*}{Method}} & \multicolumn{3}{c|}{Error (lower is better)} & \multicolumn{3}{c|}{Accuracy (higher is better)} \\ \cline{3-8}
				\multicolumn{2}{|c|}{}  &   rel      & log10    & rms     & $\delta<1.25$      & $\delta<1.25^2$     & $\delta<1.25^3$     \\
				\hline
				\multicolumn{2}{|l|}{Saxena \etal \protect\protect\cite{saxena2009make3d}}    &   0.349   &  -     &   1.214    &  0.447   &   0.745   &   0.897  \\
				\multicolumn{2}{|l|}{Liu \etal \protect\protect\cite{liu2014discrete}}                &    0.335    &   0.127    &  1.06    &   -   &   -  &   -  \\
				
				\multicolumn{2}{|l|}{Karsch \etal \protect\protect\cite{karsch2014depth}}  &        0.35 &  0.131     &  1.20    &  -      &  -     &   -        \\
				
				\multicolumn{2}{|l|}{Ladicky \etal \protect\protect\cite{ladicky2014pulling}}   &  -      &  -     &   -    &   0.542     &   0.829    &  0.941     \\
				\multicolumn{2}{|l|}{Zhou \etal \protect\protect\cite{zhuo2015indoor}} &   0.305     &   0.122     &  1.04  &  0.525    &  0.838   &   0.962    \\
				\multicolumn{2}{|l|}{Liu \etal \protect\protect\cite{liu2016learning}}   &  0.213     &   0.087    &  0.759   &   0.650     &  0.906     &   0.976    \\
				\multicolumn{2}{|l|}{Roi and Todorovic \protect\protect\cite{roy2016monocular}}    &  0.187      &  0.078     &  0.744     &   -     &  -     &   -    \\
				\multicolumn{2}{|l|}{Eigen \etal \protect\protect\cite{eigen2014depth}} & 0.215 & - & 0.907 & 0.611 & 0.887 & 0.971 \\
				\multicolumn{2}{|l|}{Eigen and Fergus \protect\protect\cite{eigen2015predicting}} & 0.158 & - & 0.641 & 0.769 & 0.950 & \textbf{0.988} \\
				\multicolumn{2}{|l|}{Laina \etal \protect\protect\cite{laina2016deeper}} & 0.129 & 0.056 & 0.583 & 0.801 & 0.950 & 0.986 \\
				\multicolumn{2}{|l|}{Xu \etal \protect\protect\cite{xu2017multi}}  &  0.139      &  0.063  &  0.609     &  0.793      &  0.948     &  0.984     \\
				\multicolumn{2}{|l|}{Xu and Wang \protect\protect\cite{xu2018structured}} & 0.121 & 0.052 & 0.586 & 0.811 & \textbf{0.954} & 0.987 \\
				\multicolumn{2}{|l|}{Joint HCRF \protect\protect\cite{wang2015towards}} & 0.220 & 0.094 & 0.745 & 0.605 & 0.890 & 0.970 \\
				\multicolumn{2}{|l|}{Jafari \etal \protect\protect\cite{jafari2017analyzing}} & 0.157 & 0.068 & 0.673 & 0.762 & 0.948 & \textbf{0.988} \\
				\multicolumn{2}{|l|}{PAD-Net \protect\protect\cite{xu2018pad}} & \textbf{0.120} & 0.055 & \textbf{0.582} & 0.817 & \textbf{0.954} & 0.987 \\
				\hline
				\multicolumn{2}{|l|}{Ours}                        &   0.122     &  \textbf{0.051}     &  \textbf{0.582}     &   \textbf{0.819}     &    0.953   &    \textbf{0.988}   \\ \hline
		\end{tabular}}
	\end{center}
\label{depthExperiment}
\vspace{-0.4cm}
\end{table}

\begin{figure*}
	\centering
	\includegraphics[width=0.8\linewidth]{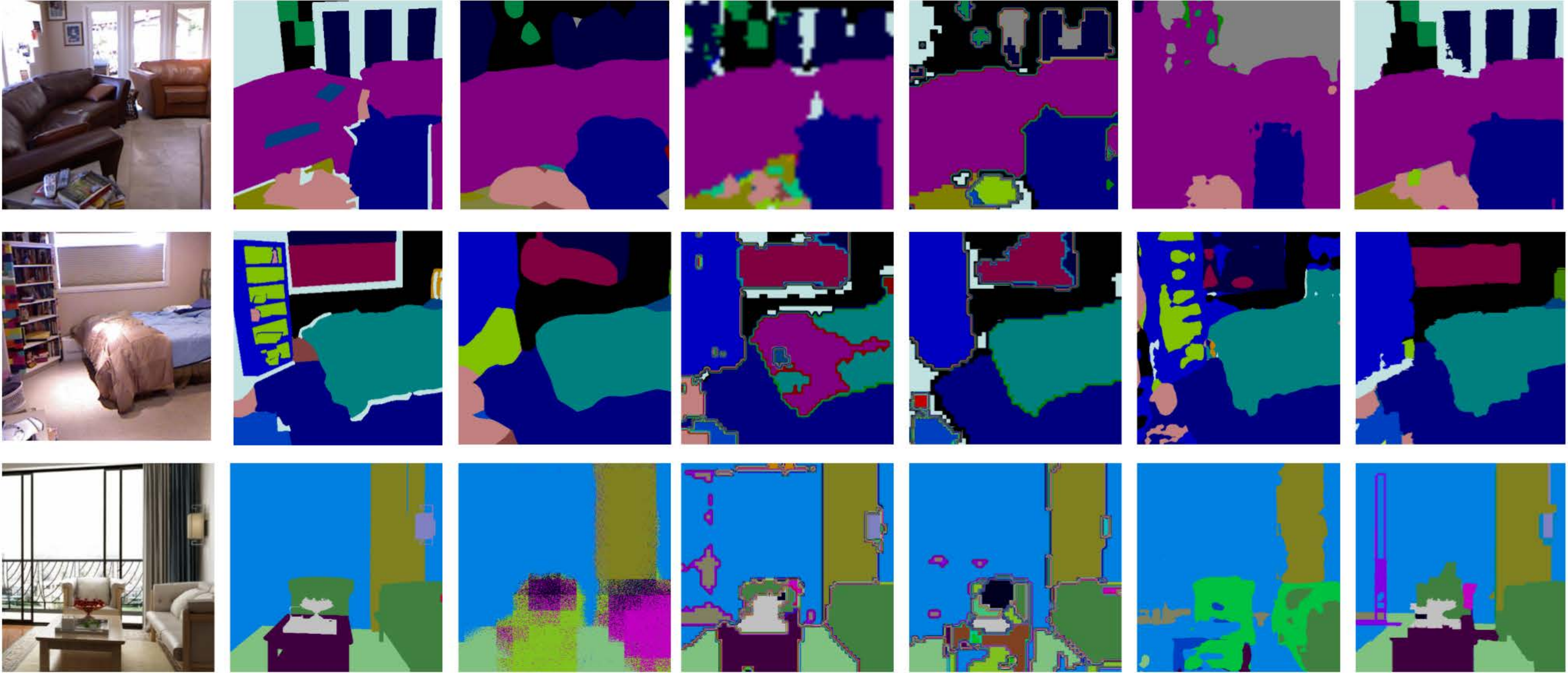}
	\caption{Semantic segmentation results on NYUv2 dataset (top two rows) and our dataset (bottom row).
	 From left to right are the input RGB images, the ground-truths and the results estimated by FCN \protect\protect\cite{long2015fully}, Chen \etal  \protect\protect\cite{chen2018deeplab}, Li \etal \protect\protect\cite{li2016lstm}, Zhao \etal \protect\protect\cite{zhao2018psanet} and our method.}	
	\label{fig:semantic_merge}
\vspace{-0.2cm}
\end{figure*}


\subsection{Semantic Segmentation}
\begin{figure*}[htbp]
	\centering
	\includegraphics[width=0.9\linewidth]{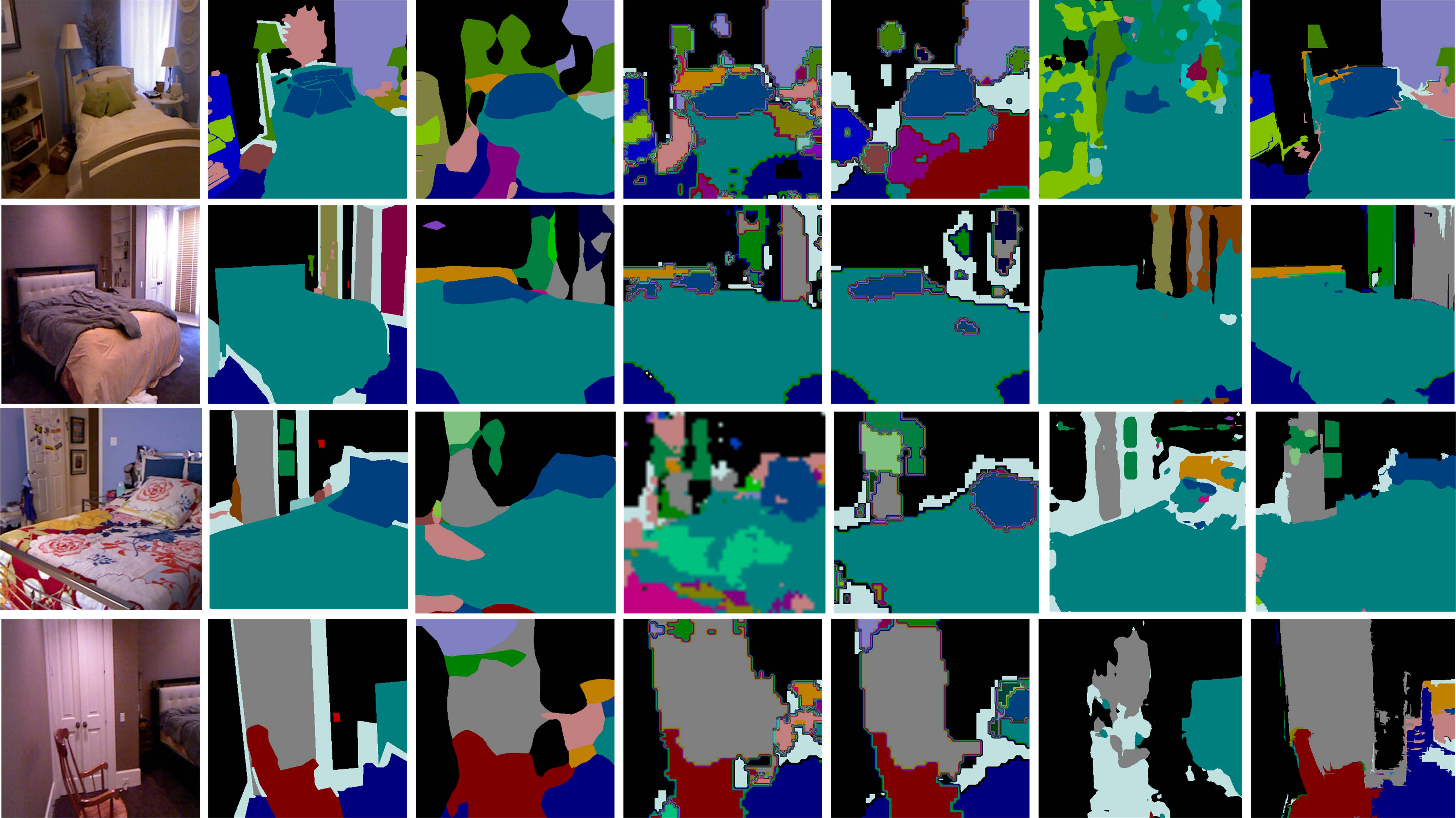}
	\caption{Semantic segmentation results on NYUv2 dataset \protect\cite{Silberman:ECCV12}. From left to right are the input RGB images, the ground-truths and the results estimated by FCN \protect\cite{long2015fully}, Chen \etal \protect\cite{chen2018deeplab}, Li \etal \protect\cite{li2016lstm}, Zhao \etal \protect\cite{zhao2018psanet} and our method.}
	\label{fig:semantic_nyu}
\end{figure*}

\begin{figure*}[htbp]
	\centering
	\includegraphics[width=0.9\linewidth]{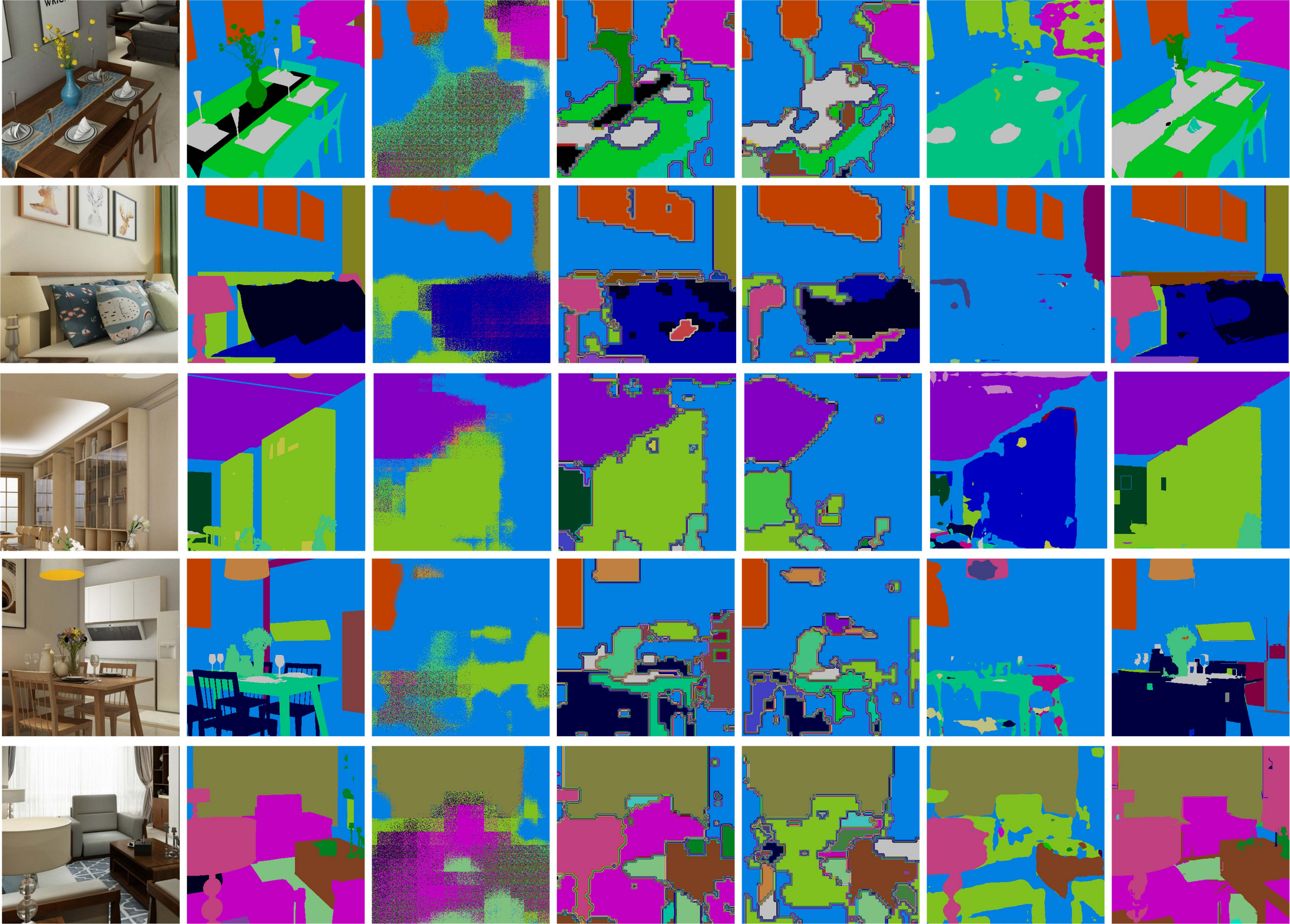}
	\caption{Semantic segmentation results on our dataset. From left to right are the input RGB images, the ground-truths and the results estimated by FCN \protect\cite{long2015fully}, Chen \etal \protect\cite{chen2018deeplab}, Li \etal \protect\cite{li2016lstm}, Zhao \etal \protect\cite{zhao2018psanet} and our method.}
	\label{fig:semantic_ourdataset}
\end{figure*}

To evaluate the performance of semantic segmentation, we use NYUv2-40 dataset \protect\cite{long2015fully} in which all objects in the NYUv2 dataset \protect\cite{Silberman:ECCV12} are divided into 40 categories. We use the same training and testing data as other methods and adopt three metrics in percentage ($\%$): pixel accuracy, mean accuracy, and Intersection over Union (IoU). As shown in Table \ref{SemanticTable}, our inferred semantic segmentation results outperform those state-of-the-art methods. We also quantitatively evaluate some recent work that provide source code on our IterNet RGB-D dataset in Table \ref{SemanticComp}. It can be seen that our method also achieves the best performance. Figure \ref{fig:semantic_merge} presents some visual comparison results on NYUv2-40 dataset and our dataset mapped into 87 categories. Being consistent with the quantitative results in Table \ref{SemanticTable} and Table \ref{SemanticComp}, our approach generates more accurate semantic segmentation results on both real dataset (NYUv2) and synthetic dataset (IterNet RGB-D) than state-of-the-art methods. More qualitative comparison results for semantic segmentation are depicted in Figure \ref{fig:semantic_nyu} and Figure~\ref{fig:semantic_ourdataset}. It can be observed that our approach generates more accurate semantic segmentation on both real dataset (NYUv2) and synthetic dataset (IterNet RGB-D) than other four competing methods.

\subsection{Multi-view Reconstruction}


\begin{table}
\caption{Quantitative evaluation for depth estimation on our dataset.}
	\renewcommand{\arraystretch}{1.3}
	\small
	\begin{center}
		\resizebox{1.0\linewidth}{!}{
			\begin{tabular}{|c|c|c|c|c|c|c|c|}
				\hline
\multicolumn{2}{|c|}{\multirow{2}{*}{Method}} & \multicolumn{3}{c|}{Error (lower is better)} & \multicolumn{3}{c|}{Accuracy (higher is better)} \\ \cline{3-8}
				\multicolumn{2}{|c|}{} & rel	&  log10	&	rms  &  $\delta<1.15$  &   $\delta<1.15^2$ &  $\delta<1.15^3$\\
				\hline
				\multicolumn{2}{|l|}{Eigen \etal \protect\protect\cite{eigen2014depth}} & 0.948	& 0.285	& 4.711 & 0.054	& 0.205	& 0.492	 \\
				\multicolumn{2}{|l|}{Laina \etal \protect\protect\cite{laina2016deeper}} & 0.404	& 0.235	& 3.433	& 0.102	& 0.310	& 0.581 \\
				\multicolumn{2}{|l|}{Xu \etal \protect\protect\cite{xu2017multi}} &  0.175	& 0.089	& 1.010	& 0.435	& 0.700	& 0.907 \\
				\multicolumn{2}{|l|}{Xu and Wang \protect\protect\cite{xu2018structured}} & 0.151	& 0.067	& 0.620	 & 0.536	& 0.817	& 0.975\\
				\hline
				\multicolumn{2}{|l|}{Ours} &  \textbf{0.136} 	& \textbf{0.062} &  \textbf{0.507}	& \textbf{0.568}	& \textbf{0.918}	& \textbf{0.982} \\							
				\hline
		\end{tabular}}
	\end{center}
	\label{depthComp}
	\vspace{-0.2cm}
\end{table}

\begin{table}
	\caption{Quantitative evaluation for semantic segmentation on the NYUv2-40 dataset.}
	\renewcommand{\arraystretch}{1.3}
	\scriptsize
	\begin{center}
		\resizebox{1.0\linewidth}{!}{
			\begin{tabular}{|l|c|c|c|}
				\hline
				Method & Pixel Accuracy  & Mean Accuracy  & IoU  \\
				\hline
				Deng \etal \protect\protect\cite{deng2015semantic}  & 63.8 & 31.5 & - \\
				FCN \protect\protect\cite{long2015fully}	  & 60.0 & 42.2  & 29.2 \\
				FCN-HHA \protect\protect\cite{long2015fully}  & 65.4 & 46.1 & 34.0 \\
				Eigen \etal \protect\protect\cite{eigen2015predicting}	   & 65.6  & 45.1  & 34.1  \\
				Lin \etal \protect\protect\cite{lin2016efficient}  & 70.0 & 53.6 & 40.6 \\				
				RefineNet\protect\protect\cite{lin2017refinenet}  & 73.6 & 58.9 & 46.5 \\
				Kong \etal \protect\protect\cite{kong2017recurrent} & 72.1 & - & 44.5 \\
				Saxena \etal \protect\protect\cite{saxena2009make3d}  & - & 55.7 & 43.1 \\
				Gupta \etal \protect\protect\cite{gupta2014learning}  & 60.3 & - & 28.6 \\
				Mousavian \etal \protect\protect\cite{mousavian2016joint}  & 68.6  & 52.3  & 39.2 \\
				
				\hline
				Ours 	&	\textbf{74.3}	 &  \textbf{59.4}	&	\textbf{48.7} \\
				\hline
		\end{tabular}}
	\end{center}
	\label{SemanticTable}
\vspace{-0.2cm}
\end{table}

\begin{table}
\caption{Quantitative evaluation for semantic segmentation on our dataset.}
	\renewcommand{\arraystretch}{1.3}
	\footnotesize
	\begin{center}
		\begin{tabular}{|l|c|c|c|}
			\hline
			Method & Pixel Accuracy	&  Mean Accuracy	&	IoU \\
			\hline
			FCN \protect\protect\cite{long2015fully} & 47.07	& 33.76	& 24.63	 \\
			Chen \etal \protect\protect\cite{chen2018deeplab} & 66.28	& 67.98	& 53.90	 \\
			Li \etal \protect\protect\cite{li2016lstm} &  61.97	& 46.93	& 40.46	 \\
			Zhao \etal \protect\protect\cite{zhao2018psanet} & 74.82		& 	72.36	& 60.91	 \\
			\hline
			Ours &  \textbf{75.54} 	& \textbf{74.49} &  \textbf{63.98}	 \\							
			\hline
		\end{tabular}
	\end{center}
	\label{SemanticComp}
\vspace{-0.4cm}
\end{table}

\begin{figure*}
	\centering
		\includegraphics[width=0.85\linewidth]{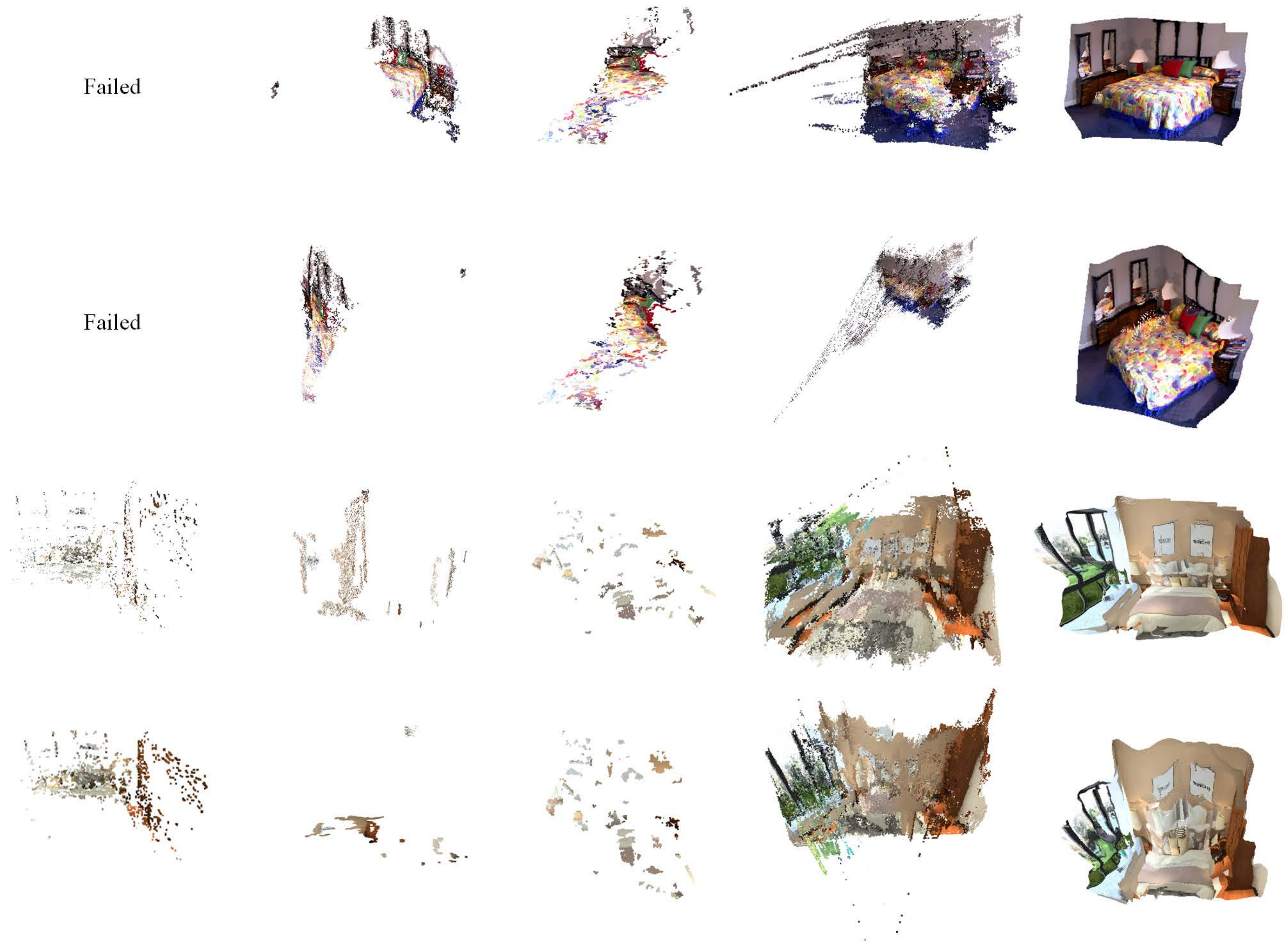}
		\caption{Comparison of scene reconstruction results of different methods on NYUv2 dataset (top two rows) and our dataset (bottom two rows). From left to right are the results of COLMAP \protect\protect\cite{schoenberger2016sfm,schoenberger2016mvs}, PMVS2 \protect\protect\cite{Furukawa2010Accurate}, OpenMVS \protect\protect\cite{OpenMVS}, DeepMVS \protect\protect\cite{DeepMVS} and our method.}
	\label{fig:best}
\vspace{-0.2cm}
\end{figure*}

\begin{figure*}
	\centering
		\includegraphics[width=0.85\linewidth]{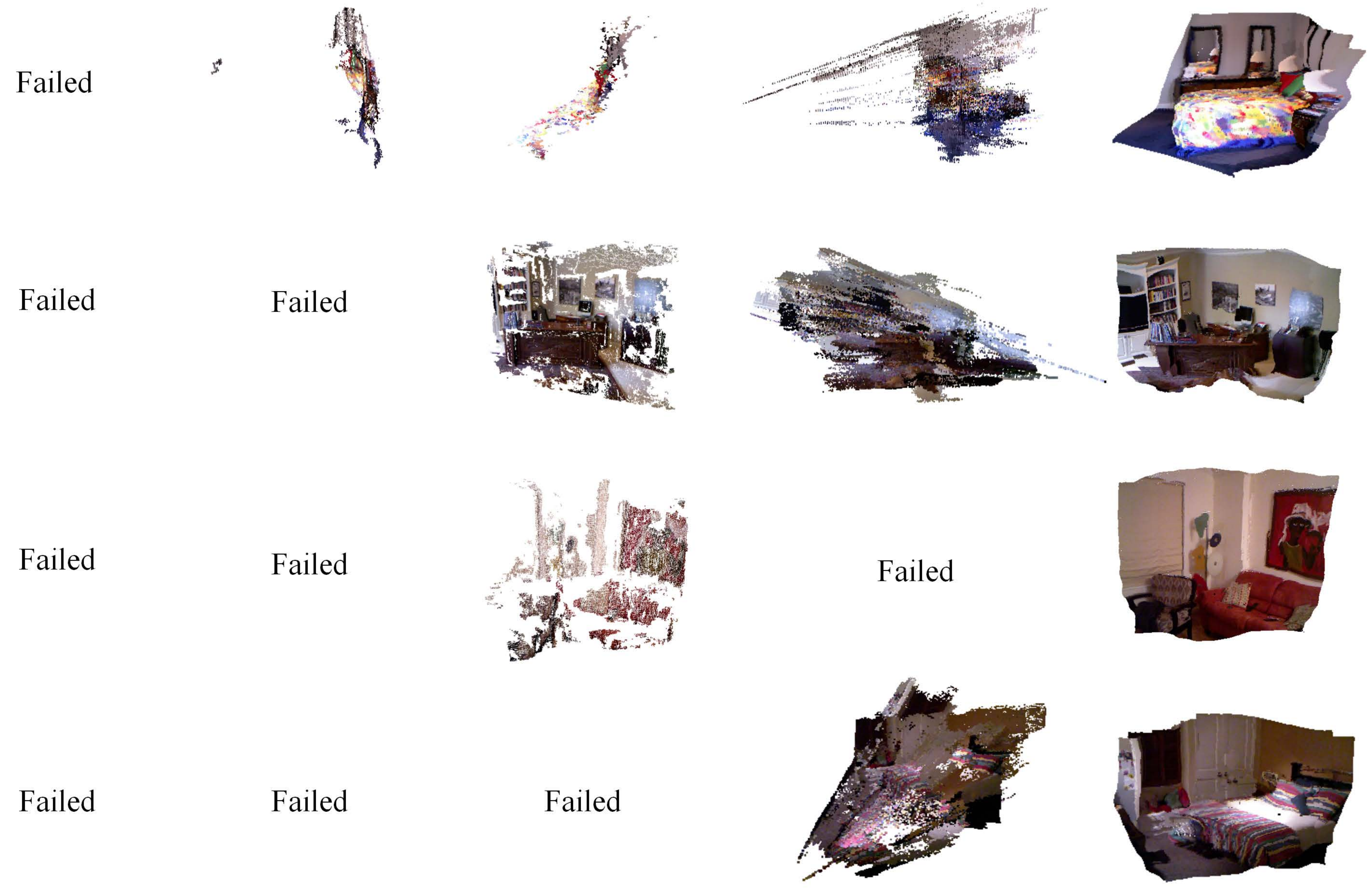}
		\caption{Comparison of multi-view reconstruction results of different methods on NYUv2 dataset \protect\protect\cite{Silberman:ECCV12}. From left to right are the results of COLMAP \protect\protect\cite{schoenberger2016sfm,schoenberger2016mvs}, PMVS2 \protect\protect\cite{Furukawa2010Accurate}, OpenMVS \protect\protect\cite{OpenMVS}, DeepMVS \protect\protect\cite{DeepMVS} and our method.}
	\label{fig:modelComp}
\end{figure*}

\begin{figure*}
	\centering
		\includegraphics[width=0.9\linewidth]{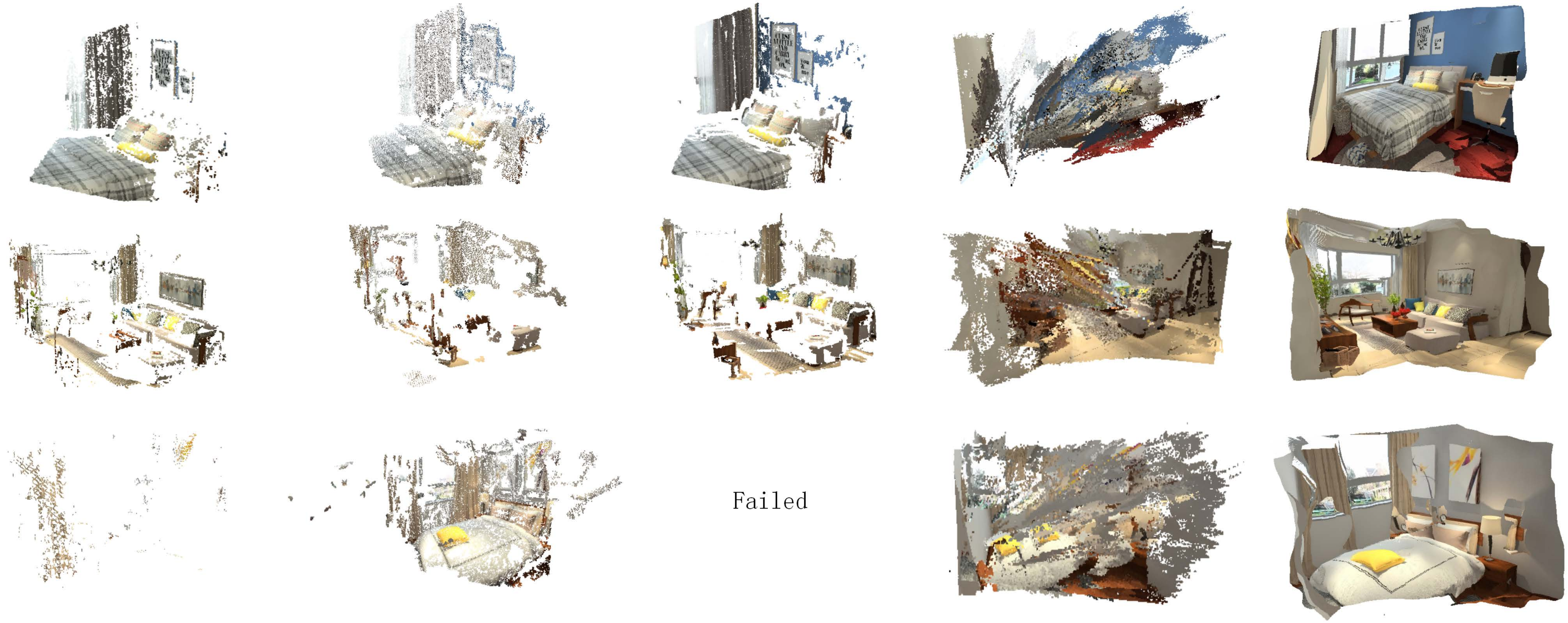}
		\caption{Comparison of scene reconstruction results of different methods on our dataset. From left to right are the results of COLMAP \protect\protect\cite{schoenberger2016sfm,schoenberger2016mvs}, PMVS2 \protect\protect\cite{Furukawa2010Accurate}, OpenMVS \protect\protect\cite{OpenMVS}, DeepMVS \protect\protect\cite{DeepMVS} and our method.}
	\label{fig:modelComp2}
\end{figure*}

\begin{figure*}
	\centering
		\includegraphics[width=0.95\linewidth]{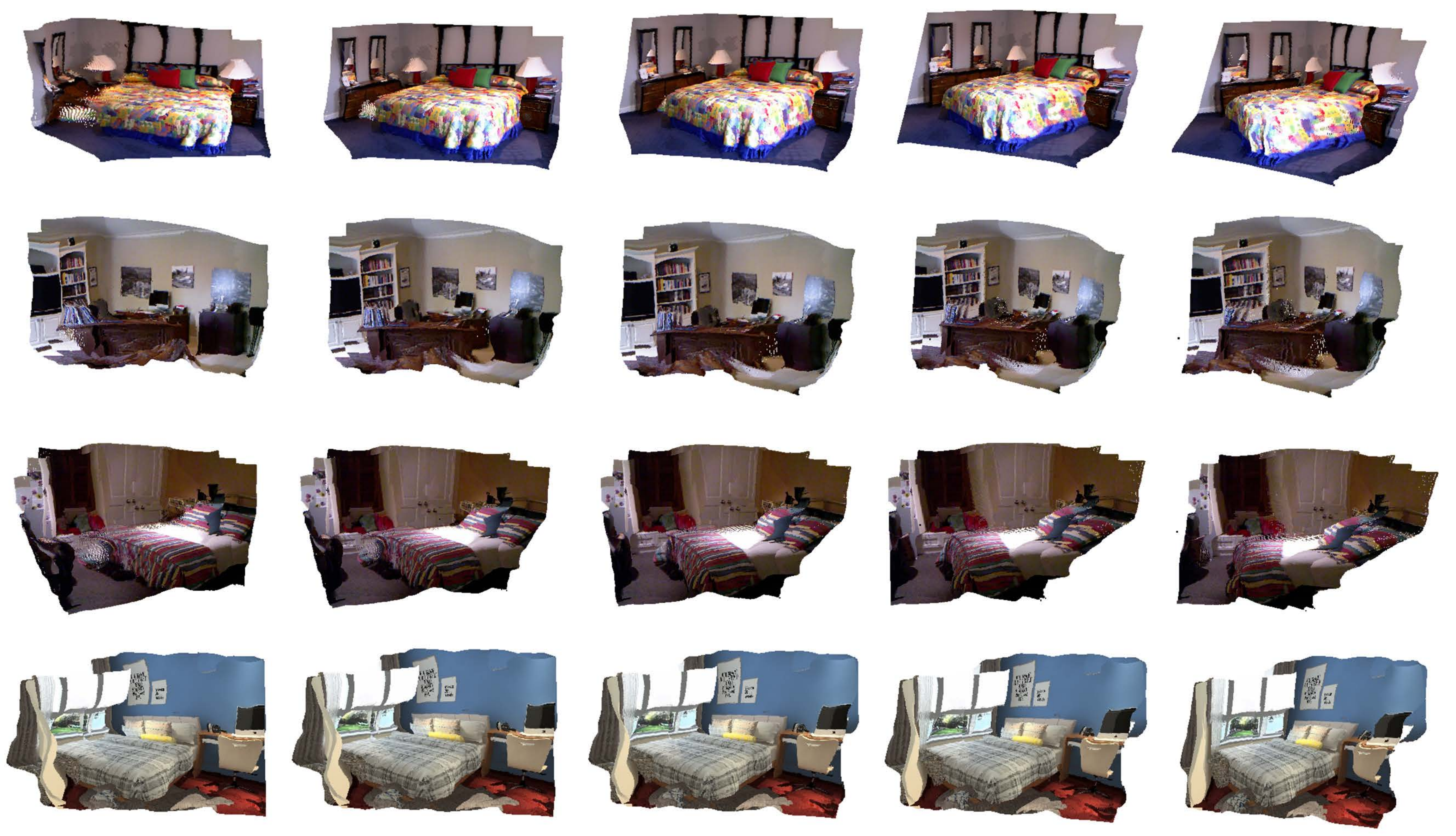}
		\caption{Our reconstructed models on NYUv2 dataset \protect\protect\cite{Silberman:ECCV12} and our dataset presented from five different views as illustrated for each scene.}
	\label{fig:mv}
\end{figure*}

In Figure \ref{fig:best}, we evaluate multi-view 3D reconstruction performance of the proposed method on NYUv2 dataset \protect\cite{Silberman:ECCV12} and our dataset using three wide-baseline views, compared with four state-of-the-art multi-view stereo methods: COLMAP \protect\cite{schoenberger2016sfm,schoenberger2016mvs},  PMVS2 \protect\cite{Furukawa2010Accurate}, OpenMVS \protect\cite{OpenMVS} and DeepMVS \protect\cite{DeepMVS}. We obtain the sparse views for NYUv2 dataset by selecting 1 frame per 30-40 frames, and use the camera parameters estimated by COLMAP \protect\cite{schoenberger2016sfm} for OpenMVS \protect\cite{OpenMVS}, PMVS2 \protect\cite{Furukawa2010Accurate} and DeepMVS \protect\cite{DeepMVS}. As shown in Figure \ref{fig:best}, COLMAP \protect\cite{schoenberger2016sfm,schoenberger2016mvs} fails to generate meaningful results on NYUv2 dataset from sparse views. We can see obviously wrong points for PMVS2 \protect\cite{Furukawa2010Accurate} and OpenMVS \protect\cite{OpenMVS}: some points gather together from side view and top view on NYUv2 dataset. Moreover, their obtained point clouds are too sparse to provide acceptable results by linear interpolation. DeepMVS reconstructs more points compared with the traditional methods, but the reconstructed model contains a lot of noise and outliers. On the contrary, our method achieves the best results for sparse multi-view reconstruction by considering 7-D information (geometry, photometry and semantics) and using joint global and local registration. More results on NYUv2 dataset \protect\cite{Silberman:ECCV12} and our dataset using three or four sparse views are given in Figure \ref{fig:modelComp} and Figure \ref{fig:modelComp2}, respectively. It can be seen that the multi-view stereo method in COLMAP \protect\cite{schoenberger2016mvs} fails to generate 3D point clouds, and the point clouds reconstructed by OpenMVS \protect\cite{OpenMVS} and PMVS2 \protect\cite{Furukawa2010Accurate} lack sufficient density and completeness. Although DeepMVS \protect\cite{DeepMVS} achieves dense reconstruction, the reconstructed model contains many wrong points. In contrast, our method achieves accurate and complete reconstruction from sparse views. Because COLMAP \protect\cite{schoenberger2016mvs} fails for most scenes in NYUv2 dataset \protect\cite{Silberman:ECCV12}, we give quantitative evaluation on our dataset in Table \ref{tab:mvsEvaluation}. We use two indicators to evaluate the results of MVS reconstruction: accuracy and completeness. Accuracy represents the average distance between the points on reconstructed model and the nearest points on the ground-truth model. Completeness measures the percentage of the points on the ground-truth model that can find corresponding points on the reconstructed model within a certain distance threshold (0.1).
We generate the 3D ground-truth model by fusing multi-view ground-truth depth point clouds using ICP. As shown in Table \ref{tab:mvsEvaluation}, our method achieves the most complete reconstruction and meanwhile ensures the accuracy. Although traditional multi-view stereo methods \protect\cite{schoenberger2016mvs,Furukawa2010Accurate,OpenMVS} have higher accuracy, their reconstructed points are too sparse to provide acceptable results by linear interpolation. Figure \ref{fig:mv} shows our reconstructed models on NYUv2 dataset \protect\cite{Silberman:ECCV12} and our dataset presented from five different views.

\begin{table}[!h]
\caption{Quantitative evaluation for multi-view reconstruction.}
	\renewcommand{\arraystretch}{1.3}
	\small
	\centering
	\begin{tabular}{|c| c| c|}
		\hline
		{\multirow{2}{*}{\textbf{Method}}} & \textbf{Accuracy} & \textbf{Completeness}  \\
& (lower is better) & (higher is better) \\
		\hline
		COLMAP \protect\protect\cite{schoenberger2016mvs}  & 3.74 & 2.33\% \\
		\hline
		PMVS2 \protect\protect\cite{Furukawa2010Accurate} & 3.71 & 1.83\% \\
		\hline
		OpenMVS \protect\protect\cite{OpenMVS} & 3.68 & 1.25\%  \\
		\hline
		DeepMVS \protect\protect\cite{DeepMVS} & 21.49 & 12.47\% \\
		\hline
		Ours & 17.72 & 31.55\% \\
		\hline
	\end{tabular}
	\label{tab:mvsEvaluation}
\end{table}

\section{Conclusions}
In this paper, we solve a challenging problem: reconstructing and understanding indoor 3D scenes based on several color images captured from uncalibrated sparse views. We propose IterNet, a novel iterative network to jointly estimate depth map and semantic segmentation from a single color image, and a joint global and local registration method to reconstruct indoor 3D scenes from sparse views. We also introduce and make available IterNet RGB-D dataset, a new dataset that simultaneously provides high-resolution photorealistic RGB images, accurate depth maps, and pixel-level semantic labels for thousand of layouts. Experimental results on both public datasets and our dataset demonstrate that our method achieves the best results on depth estimation, semantic segmentation and multi-view reconstruction, compared with state-of-the-art methods.

\bibliographystyle{ieee}
\bibliography{ref}

\end{document}